%% file: main.tex
\documentclass{article}

\PassOptionsToPackage{sort&compress}{natbib}
\usepackage{iclr2027_conference,times}
\iclrfinalcopy
\setcitestyle{numbers,square,citesep={,}}
\usepackage[utf8]{inputenc} % allow utf-8 input
\usepackage[T1]{fontenc}    % use 8-bit T1 fonts
\usepackage{hyperref}       % hyperlinks
\usepackage{url}            % simple URL typesetting
\usepackage{booktabs}       % professional-quality tables
\usepackage{amsfonts}       % blackboard math symbols
\usepackage{nicefrac}       % compact symbols for 1/2, etc.
\usepackage{microtype}      % microtypography
\usepackage{xcolor}         % colors

\usepackage{graphicx}
\usepackage{pifont}
\usepackage{caption}
\usepackage{makecell}
\usepackage{wrapfig}
\usepackage{multirow}
\usepackage{tabularx}
\newcolumntype{Y}{>{\centering\arraybackslash}X}
\usepackage{enumitem}
\usepackage{needspace}
\usepackage{amsmath}
\DeclareMathOperator*{\argmin}{arg\,min}
\hypersetup{
    colorlinks=true,  % Removes the borders around links and colors the text instead
    citecolor=cyan,   % Sets the specific color for citation numbers
    linkcolor=cyan,  % Sets the color for internal links (like the table of contents)
    urlcolor=cyan,    % Sets the color for web links
    pdftitle={Grounded World Model: Latent Planning with Language Goals},
    pdfauthor={Quanyi Li, Lan Feng, Haonan Zhang, Wuyang Li, Letian Wang, Alexandre Alahi, Harold Soh},
    pdfsubject={Preprint}
}

\newcommand{\wiserOracle}{93\%}
\newcommand{\wiserOracleTrain}{90\%}
\newcommand{\wiserGwm}{87\%}
\newcommand{\wiserGwmTrain}{92\%}
\newcommand{\wiserNoPred}{9\%}
\newcommand{\wiserDreamDojo}{17\%}
\newcommand{\wiserVlaTrain}{90\%}
\newcommand{\wiserVlaTest}{22\%}
\newcommand{\wiserVlaBest}{47\%}
\newcommand{\wiserTwoB}{33\%}
\newcommand{\wiserPE}{0\%}
\newcommand{\wiserAC}{24\%}
\newcommand{\wiserXarm}{83\%}
\newcommand{\wiserHalfD}{72\%}
\newcommand{\wiserZdirect}{61\%}
\newcommand{\wiserZdirectTrain}{92\%}
\newcommand{\wiserWemm}{76\%}
\newcommand{\wiserWemmTrain}{92\%}
\newcommand{\wiserWemmOracle}{88\%}
\newcommand{\wiserWemmOracleTrain}{94\%}
\newcommand{\simGwm}{70/70}
\newcommand{\simCamOne}{65/70}
\newcommand{\simTiptop}{94\%}
\newcommand{\simTiptopFrac}{66/70}
\newcommand{\simPiFive}{7/70}
\newcommand{\simPiFiveRelaxed}{53\%}
\newcommand{\simPiFiveRelaxedFrac}{37/70}
\newcommand{\simVjepa}{25/70}
\newcommand{\cemGrid}{4/4}
\newcommand{\cemBest}{4/4}
\newcommand{\drawerSel}{5/6}

\newcommand{\pushTotal}{153/300}
\newcommand{\pushDir}{153/153}
\newcommand{\hwGwmTotal}{55/60}
\newcommand{\hwGwmSceneOne}{36/40}
\newcommand{\hwGwmSceneTwo}{19/20}
\newcommand{\hwTiptopSceneOne}{37/40}
\newcommand{\hwTiptopSceneTwo}{15/20}
\newcommand{\hwTiptopTotal}{52/60}
\newcommand{\hwShoulder}{34/40}
\newcommand{\hwGeminiMedian}{6.0}
\newcommand{\hwGeminiMin}{4.7}
\newcommand{\hwGeminiMax}{10.0}

\title{Grounded World Model: Latent Planning with Language Goals}

\author{%
  Quanyi Li\thanks{Equal contribution.} \\
  NUS \\
  \And
  Lan Feng\footnotemark[1] \\
  EPFL \\
  \And
  Haonan Zhang \\
  Beihang University \\
  \And
  Wuyang Li \\
  EPFL \\
  \AND
  Letian Wang \\
  University of Toronto \\
  \And
  Alexandre Alahi \\
  EPFL \\
  \And
  Harold Soh \\
  NUS \\
}

\begin{document}
\raggedbottom
\maketitle
\lhead{Preprint}

%===============================================================================

\begin{abstract}
World models such as DINO-WM and LeWM specify the goal with an image, which is difficult to obtain in advance for novel tasks.
We present the Grounded World Model (GWM), a latent world model that enables zero-shot planning in the real world from language goals alone.
Given a candidate action sequence and the current observation, GWM predicts the future in the visual space of a pretrained video-language embedding model.
The frozen readout of this embedding model maps this imagined future and the task description into the same embedding space, where their negative cosine similarity serves as the planning cost.
Training GWM requires only offline and task-agnostic video-action pairs and no language labels.
In simulated experiments on WISER, planning with GWM, which executes the candidate action of lowest cost, solves \wiserGwm\ of 288 tasks with unseen instructions and visual signals, while ten fine-tuned VLAs average \wiserVlaTest.
We then scale GWM up with real robot data, and use it for zero-shot planning in realistic simulation and real scenes.
In the IsaacSim evaluation, planning with GWM completes all 70 trials across 14 tasks that require reasoning over referring expressions, matching a modular planner grounded by a frontier VLM, while $\pi_{0.5}$ reaches \simPiFiveRelaxedFrac.
Deployed on a real Franka, the same stack completes \hwGwmTotal\ separately evaluated pick-and-place sub-tasks, comparable to the modular planner's \hwTiptopTotal, with the full system running locally on a single consumer GPU.
Project website: \url{https://quanyili.github.io/gwm-wiser/}.
\end{abstract}

\section{Introduction}
\label{sec:intro}
\begin{figure*}[h]
    \centering
    \includegraphics[width=\linewidth]{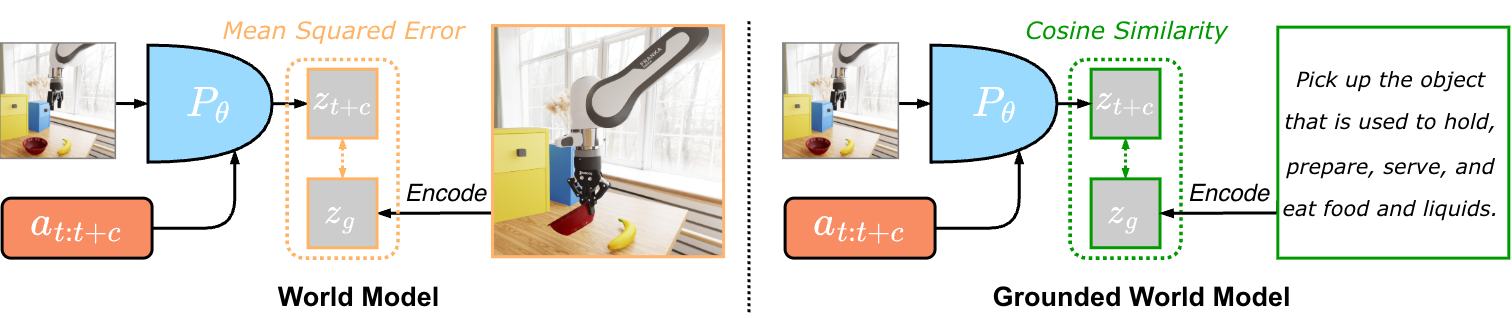}
    \caption{World Model planning with image goals and with language goals.
    \emph{Left}: the image interface, where candidate futures are imagined and compared to the goal in a pure visual latent space.
    \emph{Right}: planning with GWM, where the same imagination step runs inside the visual latent space of a video-language embedding model and a fixed readout compares the imagined future to the task description.
    }
    \label{fig:teaser}
\end{figure*}
A world model predicts how a scene evolves under an agent's actions, enabling it to optimize actions toward a goal by minimizing the cost function~\citep{ha2018recurrentworldmodelsfacilitate}.
A recent line of work has made this approach practical for visuomotor control by predicting in the representation space of a pretrained vision encoder. DINO-WM plans over frozen DINOv2 patch features~\citep{zhou2025dinowmworldmodelspretrained, oquab2024dinov2learningrobustvisual}, V-JEPA~2-AC post-trains an action-conditioned predictor on a self-supervised video encoder~\citep{assran2025vjepa2selfsupervisedvideo}, and subsequent methods refine the encoder, predictor, and training recipe~\citep{terver2026drivessuccessphysicalplanning, destrade2025valueguidedactionplanningjepa, zhang2026geoworldgeometricworldmodels, maes2026leworldmodel, sobal2026learning, zhang2026hierarchical}.
These systems share a common planning interface, in which candidate action sequences are imagined forward in latent space and the candidate whose predicted future is closest to the representation of a \emph{goal image} is executed (Fig.~\ref{fig:teaser}, left).
However, specifying the goal with images is restrictive.
A photograph of the desired end state in the target scene must be obtained before the robot can act, and a multi-stage task requires a separate photograph for each sub-goal.
For example, to perform a single pick-and-place task, V-JEPA~2-AC requires two sub-goal images in addition to the final goal image and switches among them on a fixed schedule~\citep{assran2025vjepa2selfsupervisedvideo}.
Language is an easier goal interface to specify. It is available before the robot acts, and a long-horizon task can be decomposed to sub-instructions.
But existing world models predict in a purely visual latent space, which provides no direct comparison between an imagined future and a language goal.

A straightforward fix is to align such a visual latent space with language through additional multimodal training~\citep{liu2024groundingdinomarryingdino}.
We invert this construction. Rather than aligning a world model's latent space with language, we build the world model \emph{inside a space that is already aligned with language}.
This design enables predicting task-agnostic visual future, while retaining the ability to score candidate actions against open-vocabulary instructions.
Specifically, we start from a frozen video-language embedding model trained to retrieve videos from text.
The model comprises a vision encoder that maps a video clip to a token-level latent and a multimodal readout that projects visual latents and text into a shared unit-norm embedding space, where cosine similarity measures semantic agreement.
With this base model, we build \textbf{Grounded World Model (GWM)} to predict, from the current frame and a candidate action sequence, the token-level latent of the corresponding future video clip.
At test time, the frozen readout maps this prediction and the instruction into the shared embedding space.
During planning, the predicted future of each candidate action sequence is compared with the task instruction in this space, and the planner selects the candidate action sequence that minimizes their negative cosine similarity (Fig.~\ref{fig:teaser}, right).
We call the world model \emph{grounded} because its predictions can be grounded to language by this readout, although no language label is ever used to train it.

This construction has four benefits.
First, the goal is stated in language, yet training needs only offline and task-agnostic video-action pairs, without even success filtering.
Second, the dynamics stay task-agnostic and the cost function is built in zero-shot given the instruction. One model serves all tasks.
Third, actions enter as rendered pixels, so no action tokenizer is trained and any robot with a kinematic model and camera parameters can be scored.
Fourth, the full stack runs on one local consumer GPU.
Our experiments support this design through four key findings:
\begin{itemize}[leftmargin=1em, itemsep=0.1em, topsep=0.em]
    \item \textbf{Language grounds intent in imagined futures.} On WISER, a controlled testbed that separates grounding from motor-skill generalization, planning over GWM's predicted futures solves \wiserGwm\ of 288 held-out tasks.
    % This closely approaches the \wiserOracle\ achieved by planning over ground-truth futures. 
    In contrast, ten VLAs fine-tuned from VLMs on the same demonstrations average only \wiserVlaTest. Ablation on WISER yields a robust recipe to support real-world data training.
    
    \item \textbf{The cost supports semantic grounding and contact-accurate action selection.} When trained on real robot data, GWM's cost function successfully guides a cross-entropy method (CEM)~\citep{rubinstein2004cross} search to localize four named photos on a tabletop. Furthermore, the same planner generates precise actions to push target cubes, achieving a 51\% success rate in moving the target at least 3\,cm without disturbing unrelated objects.
    
    \item \textbf{Zero-shot planning generalizes to open scenes.} In IsaacSim evaluation with realistic rendering, the GWM cost ranks language-blind trajectory proposals and the planner executes the one that matches the instruction. Planning with GWM completes \simGwm\ trials across 14 tasks requiring referring-expression reasoning, matching a modular planner (\simTiptop) grounded by Gemini-Robotics-ER~\citep{geminiroboticsteam2025geminiroboticsbringingai}, while $\pi_{0.5}$ reaches \simPiFiveRelaxed\ under a relaxed pick criterion.
    
    \item \textbf{The system is capable of zero-shot real-robot deployment.} We ask the same GWM planner to operate a Franka arm in an out-of-the-box evaluation with out-of-distribution viewpoint, novel objects, unseen instructions.
    The system completes \hwGwmTotal\ separately evaluated pick-and-place tasks, comparable to the modular planner's \hwTiptopTotal, with the entire stack running on a single NVIDIA RTX 5090 within 24\,GB of GPU memory instead of a remotely hosted frontier VLM.
\end{itemize}

\section{Related Work}
\label{sec:related}
\textbf{World models and goal specification.}
DINO-WM~\citep{zhou2025dinowmworldmodelspretrained}, V-JEPA~2-AC~\citep{assran2025vjepa2selfsupervisedvideo}, and a rapidly growing family of successors~\citep{terver2026drivessuccessphysicalplanning, destrade2025valueguidedactionplanningjepa, wang2026temporalstraighteninglatentplanning, zhang2026geoworldgeometricworldmodels, goswami2025worldmodelsleveragehuman, maes2026leworldmodel, sobal2026learning, zhang2026hierarchical, bar2025navigation} predict future latents of a frozen or jointly learned encoder and plan by minimizing a latent distance to an encoded goal image.
These methods are demonstrated in simulation and, for V-JEPA~2-AC and HWM, zero-shot on real Franka arms.
The world model remains task-agnostic because the task enters only through the cost~\citep{lecun2022path, sobal2026learning}.
Our method keeps this recipe, including a frozen encoder, reward-free offline latent prediction, and test-time optimization, while using language~\citep{lin2024learningmodelworldlanguage, du2023learninguniversalpoliciestextguided} to define the cost.

\textbf{Language as a planning cost.}
GenRL~\citep{mazzaglia2024genrl} connects a video-language embedding to a reconstruction-based world model through a learned adapter and trains policies in imagination.
FOREWARN~\citep{wu2025foresight} rolls out a world model and has a fine-tuned VLM verbalize and rank the rollouts.
VLMPC~\citep{zhao2024vlmpcvisionlanguagemodelpredictive} and semantic world models~\citep{berg2025semantic} query a VLM on generated video or VQA-style predictions.
LAGO~\citep{barbeau2026latent} learns language-to-latent-subgoal mappings from thousands of language-annotated demonstrations per environment.
These approaches learn an additional interface to the world model, use a generative VLM to evaluate candidate futures, or both.
The GWM cost compares embeddings with a single cosine after the frozen readout, with no language labels for world-model training, no generative VLM queries, and no per-environment adaptation.
Related reward models score robot videos against language for policy learning or verification, rather than for world-model planning~\citep{lee2026roborewardgeneralpurposevisionlanguagereward, liang2026robometerscalinggeneralpurposerobotic, kwok2026scalingverificationeffectivescaling, chen2026toprewardtokenprobabilitieshidden, kang2025cliprtlearninglanguageconditionedrobotic}.

\textbf{Modular VLM planners and VLAs.}
TiPToP~\citep{shen2026tiptop} grounds instructions with the proprietary VLM, Gemini-Robotics-ER~\citep{geminiroboticsteam2025geminiroboticsbringingai} and open-vocabulary segmentation, then plans with task-and-motion optimization.
It shares the geometric processing, grasp sampler, and motion planner with our real-world system and is our main comparison.
VLAs fine-tune a pretrained VLM into an end-to-end policy~\citep{brohan2023rt2visionlanguageactionmodelstransfer, black2026pi0visionlanguageactionflowmodel, intelligence2025pi05visionlanguageactionmodelopenworld, yang2025instructvlavisionlanguageactioninstructiontuning}.
Growing evidence indicates fine-tuning erodes inherited vision-language alignment, leading to poor generalization~\citep{zhai2025ignitingvlmsembodiedspace, song2025reconvla, zhang2025vlaseffectivelyinheritvlms, li2025taskreconstructionextrapolationpi0, zhou2025liberoprorobustfairevaluation, xu2025seeingactpromptingspecify, wang2026liberoxrobustnesslitmusvisionlanguageaction}.
The VLA train-test gap on WISER is consistent with this pattern.
GWM training leaves the pretrained embedding model's weights unchanged.
On WISER, planning with GWM retains most of its training success on the held-out split (\wiserGwmTrain\ to \wiserGwm) while the VLAs fall from \wiserVlaTrain\ to \wiserVlaTest.

\section{Grounded World Model}
\label{sec:method}
A latent world model for visuomotor control consists of an observation model that encodes observation $o_t$ into a latent state and a transition model that predicts the latent consequence of a candidate action sequence $a_{t:t+c}$ over horizon $c$~\citep{zhou2025dinowmworldmodelspretrained}.
A planner specifies a goal and a cost that scores predicted futures, then selects the minimum-cost candidate action sequence.
In prior works, the encoder is a self-supervised vision model like DINO and the goal is an image $o_g$, so the cost is a latent distance $\|\hat{z}_{t+c} - \mathrm{enc}(o_g)\|$ between predicted future and encoded goal.

\subsection{From a Video-Language Embedding Model to a World Model}
\label{sec:gwm}
To make world models more broadly useful, we specify the goal with language.
One could align that purely visual space with language through further multimodal training.
We invert the construction and build the world model inside a space already aligned with language, that of a video-language embedding model.
Trained for retrieval, it consists of a vision encoder $E$ that maps an image or video clip to a token-level latent, and a multimodal readout $B$ that pools such a latent, optionally with text, into a unit-norm embedding.
As a result, cosine similarity between a clip's embedding and a text's embedding measures how well the text describes the clip.

\begin{figure*}[t]
    \centering
    \includegraphics[width=\linewidth]{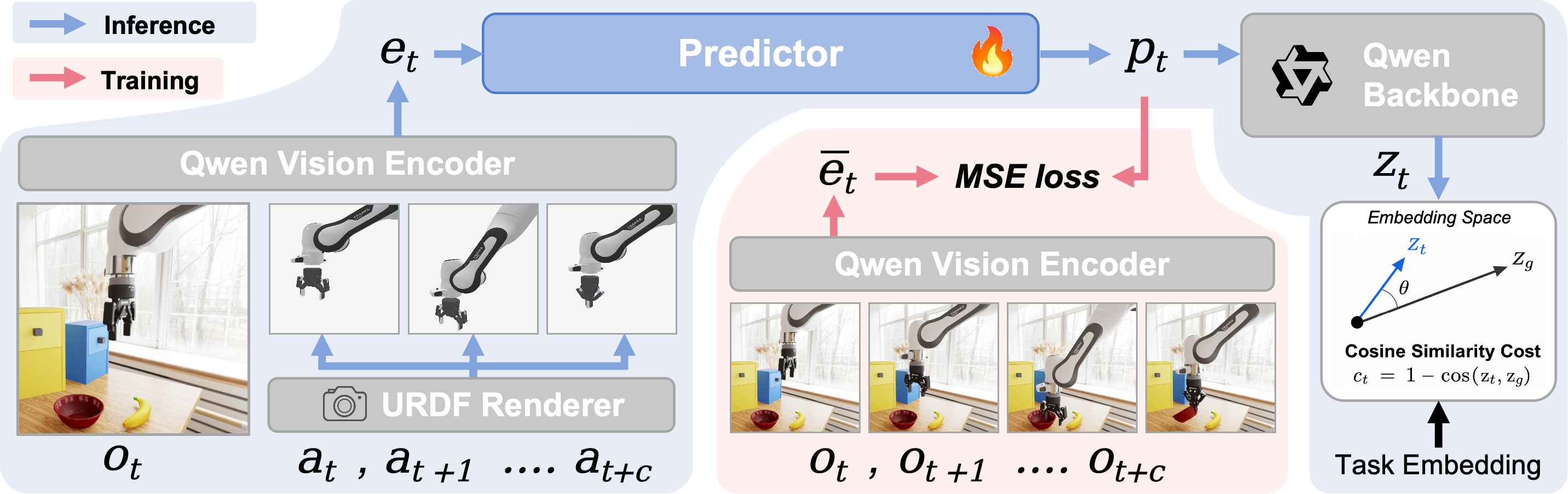}
    \caption{Training GWM and reading out its predictions for planning, shown for one candidate of a bowl pick-up task.
    \emph{Blue flow}: the candidate trajectory $a_{t:t+c}$ is rendered by $R$ into robot-only frames and encoded together with the current observation into $e_t$ by the frozen vision encoder $E$; the learned predictor $P_\theta$ outputs the imagined future latent $p_t$, which the frozen readout $B$ projects to $z_t$ for comparison with the instruction embedding $z_g$, where $1-\cos$ ranks candidates as Eq.~\ref{eq:cost} does.
    \emph{Pink flow}: during training the ground-truth future clip is encoded by the same $E$ into the target $\bar{e}_t$ of Eq.~\ref{eq:training}.
    }
    \label{fig:method_gwm}
\end{figure*}

\textbf{Definition.}
A \emph{Grounded World Model} comprises the frozen vision encoder $E$ as its observation model and a learned predictor $P_\theta$ as its transition model (Fig.~\ref{fig:method_gwm}).
It maps the current frame and a candidate action sequence to the token-level latent of the corresponding future video clip.
GWM is not language-conditioned. It is language-grounded because it predicts in a visual latent state space from which a fixed pretrained readout $B$ maps imagined futures and language goals into the same embedding space.
The readout $B$ is not part of GWM and enters only through the planning cost.
Only $P_\theta$ is learned, while $E$ and $B$ are never updated, preserving the pretrained video-language alignment and world knowledge for grounding.
In practice, we instantiate $E$ and $B$ with Qwen3-VL-Embedding-8B~\citep{li2026qwen3vlembeddingqwen3vlrerankerunifiedframework} and WeMM-Embedding~\citep{wemm-embedding}.

\textbf{Observations and actions as video.}
A candidate trajectory $a_{t:t+c}$ is tokenized by a renderer $R$ into $K{-}1$ robot-only frames, a scheme we call Rendering-based Action Tokenization (RAT).
Each frame renders the robot mesh at one future joint state from the robot's URDF, using the scoring camera's parameters, namely its intrinsics and extrinsics, so the rendered arm appears where the real arm would appear in $o_t$.
Everything except the robot is left black, since how the scene responds is what $P_\theta$ has to predict.
Prepending the current frame gives a $K$-frame condition clip, encoded as $e_t = E(o_t, R(a_{t:t+c}))$.
We use $K{=}6$.
Actions enter as pixels, requiring no new tokenizer parameters, and any robot with a kinematic model and camera parameters can get its action encoded.

\textbf{Predictor.}
$P_\theta$ is trained so that its output for the condition clip matches the latent that $E$ assigns to the real $K$ frames' future clip, $p_t = P_\theta(e_t) \approx \bar{e}_t = E(o_{t:t+c})$.
Training requires only a dataset $\mathcal{D}$ of offline video-action trajectories.
Training minimizes the latent-space error of this prediction,
\begin{equation}
\theta^\ast = \argmin_\theta \, \mathbb{E}_{(o_{t:t+c},\, a_{t:t+c}) \sim \mathcal{D}} \, \big\| P_\theta\big(E(o_t, R(a_{t:t+c}))\big) - E(o_{t:t+c}) \big\|_2^2.
\label{eq:training}
\end{equation}
The loss asks $P_\theta$ only to learn the transitions observed in the offline data.
Semantic comparison with the goal is left to the fixed readout $B$.
We predict the token-level latent rather than the pooled embedding $B(\bar{e}_t)$ because the token-level target supervises every token and keeps the spatial and temporal structure of the clip.
Regressing the pooled embedding directly with the same predictor fits the training tasks but falls to \wiserZdirect\ on the held-out split (\S\ref{sec:preliminary}).

\subsection{Grounding Predicted Futures for Planning}
\label{sec:planning}
\textbf{Grounding readout.}
Because $p_t$ is predicted in $E$'s latent space, the fixed readout $B$ treats it as an encoded video and maps the imagined future of the candidate action sequence to $z_t = B(p_t)$.
The instruction $\ell$ defines the goal and is embedded by the same fixed readout as $z_g = B(\ell, E(o_t))$, with a fixed system prompt prepended to $\ell$ and the current frame $o_t$ providing visual context.
The embedding model was pretrained to align videos with their textual descriptions, while Equation~\ref{eq:training} trains the predictor to approximate the latent encoding of the real future video.
Action scoring reuses the model's pretrained text-to-video retrieval capability, so the planner searches for the action whose imagined future best matches the instruction.
We call the composition of GWM, the fixed readout, and negative cosine similarity the \textbf{GWM cost}. Once the embeddings are computed, comparison requires one inner product per candidate.

\textbf{Planning with GWM.}
A basic planner obtains $N$ candidates $a^1, \dots, a^N$ from a trajectory proposal mechanism $q(a \mid o_t)$ and executes the one of minimum GWM cost,
\begin{equation}
\begin{aligned}
\mathcal{C}(a^n \mid o_t, \ell) &= -\big\langle B\big(P_\theta(E(o_t, R(a^n_{t:t+c})))\big),\, B(\ell, E(o_t)) \big\rangle, \\[2pt]
n^\ast &= \argmin_{n \in \{1, \dots, N\}} \, \mathcal{C}(a^n \mid o_t, \ell), \qquad a^\ast = a^{n^\ast},
\end{aligned}
\label{eq:cost}
\end{equation}
with unit-normalized embeddings, so the inner product is cosine similarity and the cost is its negative.
GWM defines a task-agnostic space of imagined futures, while each language goal induces a different cost landscape over that space.
Because GWM is agnostic to $q$, the proposals can be full trajectories or target end-effector poses $T^\ast$, which an interpolator $\mathcal{I}$ expands into a joint-space trajectory $a_{t:t+c} = \mathcal{I}(T_t, T^\ast)$ from the current pose $T_t$.
In the experiments, we use three instances of $q$.
Joint-state retrieval of demonstrated trajectories serves \S\ref{sec:preliminary}.
A Gaussian over $T^\ast$, refined by CEM~\citep{rubinstein2004cross} and expanded by $\mathcal{I}$, serves \S\ref{sec:scaling}.
Structured proposals, where grasp sampling and motion planning give each object candidate trajectories, serve \S\ref{sec:droidsim} and \S\ref{sec:hardware}.

\section{Preliminary Experiments}
\label{sec:preliminary}
Planning with GWM relies on two measurable properties.
(P1) The fixed readout $B$ must map clips and instructions so their cosine ranks futures correctly in robot scenarios.
(P2) GWM must predict the future latent from the current frame and candidate actions.
We isolate and measure both properties on a simulated testbed called WISER before moving to open scenes.

\begin{figure*}[t]
  \centering
  \includegraphics[width=\linewidth]{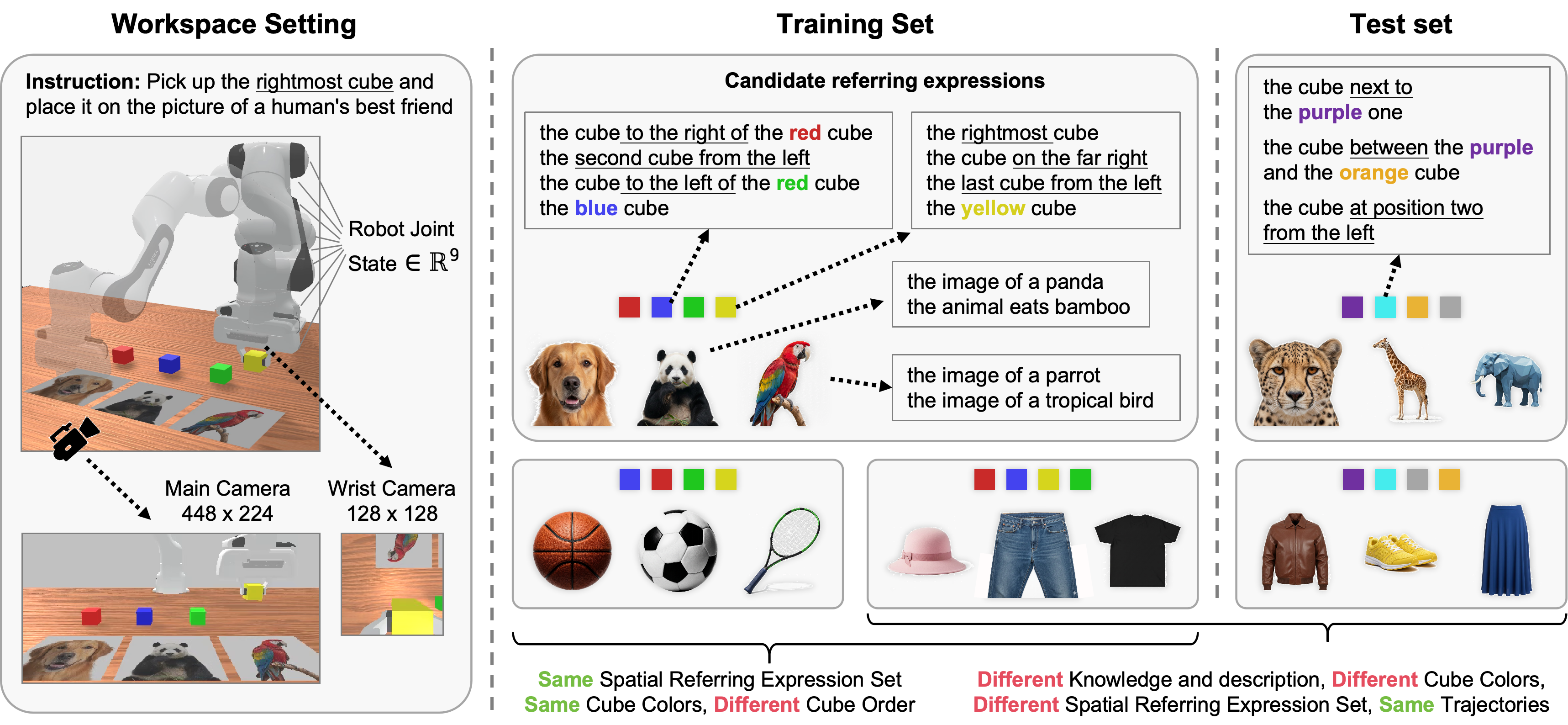}
  \caption{The WISER testbed.
  Each observation combines the instruction $\ell$, joint positions, gripper state, and camera input $o_t$.
  Each of the 24 world-knowledge categories has a training scene and a paired test scene. The two splits share the same 12 motions but have disjoint images, referring expressions, and cube colors.
  Cube order is randomized across categories, so even a fixed position, such as the second cube from the left, carries different visual and semantic cues across the splits.}
  \label{fig:wiser}
\end{figure*}

\textbf{WISER.}
Open-scene success entangles grounding with perception and action quality. 
To expose the role of grounding, we build \textbf{W}orld-knowledge \textbf{I}ntegrated \textbf{S}emantic \textbf{E}mbodied \textbf{R}easoning (WISER), which pairs 24 categories of world knowledge, including food, animals, landmarks, and numbers, with tabletop pick-and-place in ManiSkill~\citep{tao2025maniskill3gpuparallelizedrobotics}.
Each category has training and test splits (Fig.~\ref{fig:wiser}). A scene contains four cubes and three printed images, and each instruction names a cube and destination, giving 12 tasks per scene and 288 in each split.
The test visuals and instructions are disjoint from training, but the 12 motions are shared, so every test task can be solved by recalling a demonstrated motion. This separates semantic generalization from the need to learn new motor skills.

\textbf{Setup.}
We collect data on the training tasks with a scripted planner, producing 1{,}728 demonstrations, six per task.
Appendix~\ref{app:wiser} gives the collection procedure, metrics, full protocol and ablation details.
We compare planning with GWM to VLAs fine-tuned on the same demonstrations.
Planning uses closed-loop receding-horizon control. Every 20 steps, the $N{=}12$ candidates are retrieved by joint-state matching.
These trajectories are scored with Eq.~\ref{eq:cost}, and the goal query used on this testbed is given in Appendix~\ref{app:prompt}.
Retrieval supplies executable candidates, so planning tests whether the predicted futures identify the correct motion. VLAs must map the same new scenes and instructions to demonstrated actions.
The planning hyperparameters for GWM are selected with the Oracle on the training split.
The VLAs and planning with GWM are evaluated zero-shot on the test tasks.
We report the fraction of completed tasks in Table~\ref{tab:wiser} and Fig.~\ref{fig:wiser_bars}.
Detailed metrics such as target object grasp and destination reach success are reported in Appendix~\ref{app:wiser}.

\begin{wraptable}{r}{0.5\linewidth}
  % \vspace{-0.5em}
  \centering
  \footnotesize
  \caption{WISER success of GWM and its variants.}
  \label{tab:wiser}
  \setlength{\tabcolsep}{5pt}
  \begin{tabular}{lcc}
    \toprule
    Method & \makecell{Train\\Success} & \makecell{Test\\Success} \\
    \midrule
    GWM (Ours) & 0.92 & 0.87 \\
    \midrule
    GWM w/ action encoder & 0.74 & 0.24 \\
    GWM w/ pooled target & 0.92 & 0.61 \\
    GWM w/ WEMM & 0.92 & 0.76 \\
    GWM w/ xArm6 & 0.87 & 0.83 \\
    Oracle & 0.90 & 0.93 \\
    \bottomrule
  \end{tabular}
  \vspace{-0.5em}
\end{wraptable}
\textbf{The space reads intent from the futures.}
The Oracle diagnostic replaces each prediction $p^n_t$ with the ground-truth future latent $\bar{e}^n_t$, revealing what the embedding space could support under perfect prediction.
It completes \wiserOracleTrain\ of the training tasks and \wiserOracle\ of the 288 test tasks. This supports P1, showing that cosine to the instruction can rank the correct future among twelve candidates across different colors, spatial relations, and world knowledge.
The same Oracle falls to \wiserTwoB\ in the 2B sibling and to \wiserPE\ in a CLIP-style video encoder~\citep{bolya2025perceptionencoderbestvisual}.

\textbf{The future is predictable in this space.}
Planning with futures predicted by GWM completes \wiserGwm\ of the test tasks, narrowing the Oracle gap to 6 percentage points and supporting P2. Removing $P_\theta$ and scoring its input latent $e_t$ directly gives \wiserNoPred. Predicting pixels with DreamDojo~\citep{gao2026dreamdojogeneralistrobotworld}, re-encoding them with $E$, and applying the same readout gives \wiserDreamDojo. These comparisons support learning a forward predictor in latent space, which avoids synthesizing and re-encoding a future video for each candidate.
A third variant keeps the same predictor but regresses the pooled embedding $B(\bar{e}_t)$ directly and scores it without the readout. It matches GWM on the training tasks at \wiserZdirectTrain\ but falls to \wiserZdirect\ on the held-out split, favoring token-level latent prediction.
Furthermore, how actions enter the predictor is equally critical. Replacing RAT with a learned encoder over raw numeric actions yields only \wiserAC\ at test. In contrast, RAT enables the Panda-trained GWM to predict futures for visually distinct xArm6 trajectory proposals, achieving zero-shot success at \wiserXarm.

\textbf{The recipe is not tied to one embedding model.}
Replacing Qwen3-VL-Embedding-8B with WEMM~\citep{wemm-embedding}, a 9B video-language embedding model, and training the same predictor with the same recipe gives \wiserWemmTrain\ on the training tasks and \wiserWemm\ on the test tasks, while the Oracle in that space reaches \wiserWemmOracleTrain\ and \wiserWemmOracle\ (Table~\ref{tab:wiser}, Appendix~\ref{app:encoder}).
This result shows that the same recipe also works with another video-language embedding model.
Qwen3-VL-Embedding-8B performs better on WISER, so we use it for the subsequent experiments.

\begin{figure*}[t]
  \centering
  \includegraphics[width=\linewidth]{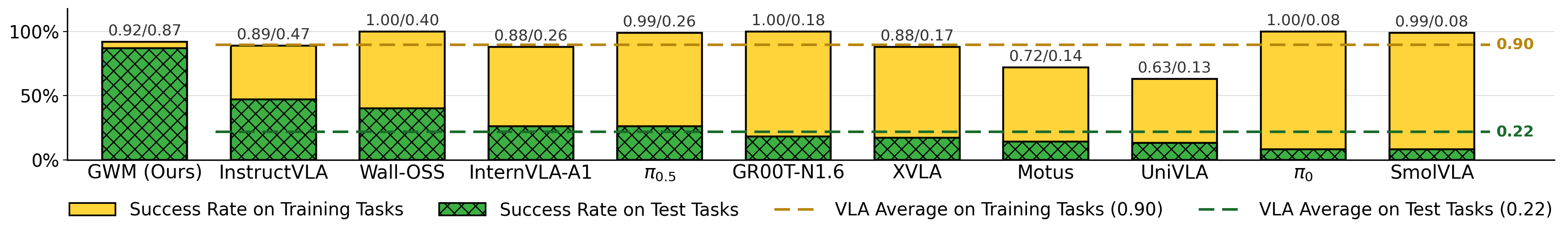}
  \caption{WISER success of planning with GWM and of ten fine-tuned VLAs on training tasks (yellow) and semantically disjoint test tasks (green). Dashed lines show the VLA averages. }
  \label{fig:wiser_bars}
\end{figure*}

\textbf{Planning in the frozen space retains semantic generalization.}
Ten VLAs, including the world-action model Motus, are fine-tuned on the same data as GWM.
Their average success falls from \wiserVlaTrain\ on training tasks to \wiserVlaTest\ on test tasks, and the best reaches \wiserVlaBest\ (Fig.~\ref{fig:wiser_bars}).
The models still perform perfect pick-and-place motions in the test scenes, but choose the wrong cube or destination. The gap suggests a poor ability to bind instruction to object despite recalling in-distribution motions.
% Fine-tuning these baselines requires 70-300 H100 GPU hours, versus 20 H100 GPU hours for GWM's predictor (Table~\ref{tab:wiser_full}).

\section{Scaling to Real-World Data}
\label{sec:scaling}

We scale the WISER-selected recipe to real-world-scale data. We train a new GWM on DROID~\citep{khazatsky2024droid} and MolmoB0T simulation data~\citep{deshpande2026molmob0t}. The resulting predictor checkpoint, paired with the frozen $E$ and $B$, is evaluated zero-shot across all subsequent experiments. Here, we define \emph{zero-shot} as deployment in an unseen environment featuring novel viewpoints, new objects, and instructions, without any target-task data, demonstrations, rewards, or fine-tuning.
Before integrating GWM into a full manipulation system, we test whether its cost landscape remains semantically grounded and spatially precise despite the shift in training data and visual realism. 

\begin{figure*}[t]
    \centering
    \includegraphics[width=\linewidth]{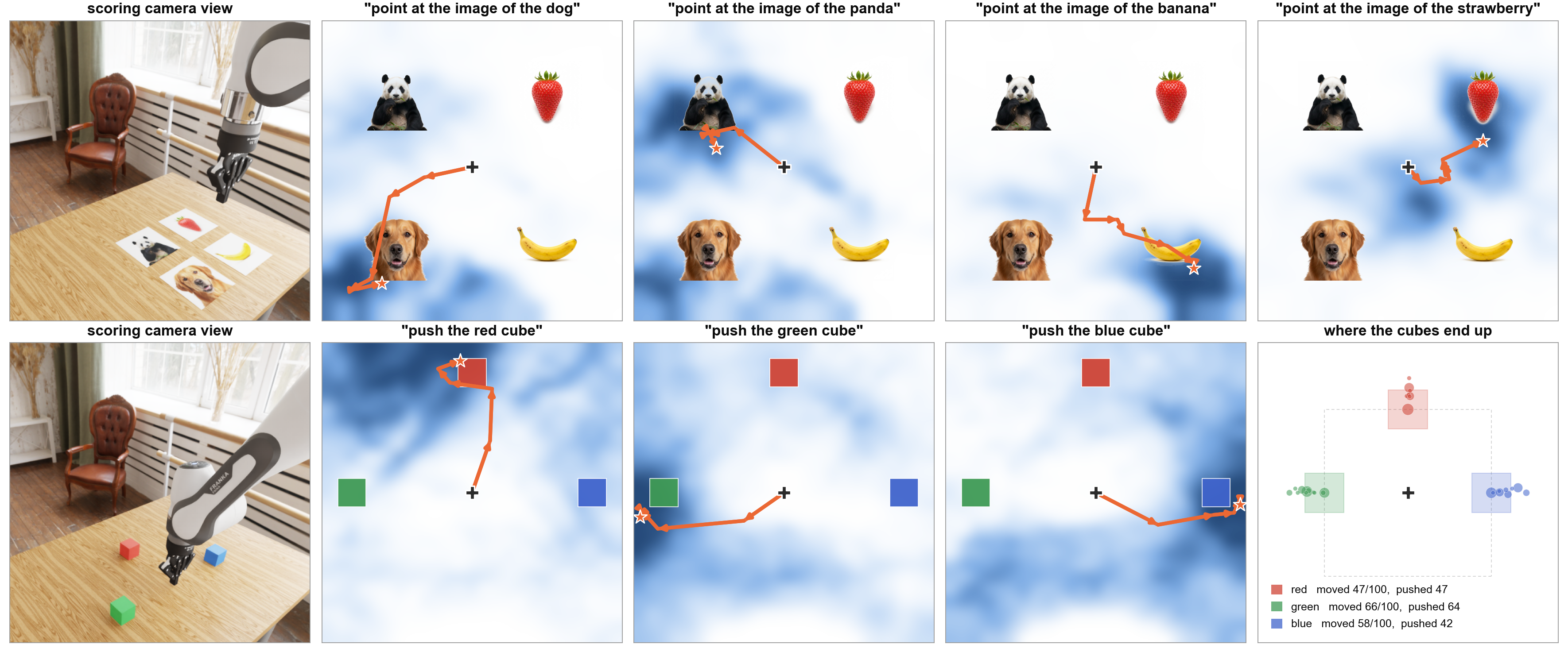}
    \caption{GWM-based scoring landscapes after scaling GWM to real-world data. Each row begins with an external scoring view.
    \emph{Top}: scores over end-effector hover positions for four pointing instructions. The orange path follows the CEM mean, and the star marks the selected final gripper pose. Darker blue denotes a higher value of the probe's scoring objective.
    \emph{Bottom}: scores over slide endpoints for three pushing instructions, followed by the named cube's final positions under each instruction. The outcome legend distinguishes ``moved'' (at least 1\,cm) from ``pushed'' (at least 3\,cm).}
    \label{fig:scaling}
\end{figure*}

\textbf{Setup.}
We build separate pointing and pushing scenes in IsaacSim. 
The pointing task evaluates target-level semantic grounding, while pushing tasks test whether the cost can distinguish trajectories that make effective contact. We use CEM~\citep{rubinstein2004cross} to test whether this cost guides search from an uninformed Gaussian over candidate endpoints.
In both, CEM searches over the tabletop endpoint $(x,y)$ of a target pose $T^\ast$, which $\mathcal{I}$ expands into a trajectory scored with GWM. Figure~\ref{fig:scaling} shows the scoring landscapes and search traces; Appendix~\ref{app:cem} gives the probe protocols.

\textbf{The cost provides robust semantic grounding.}
In the pointing task, four printed photos of a dog, a panda, a banana, and a strawberry lie on an empty table. The instruction is ``point at the image of the \{name\}.'' The visualization shows that the lowest-cost region overlaps the named photo. CEM's best sample reaches the target in \cemBest\ searches. These searches show that the cost function provides a correct semantic signal to guide the search and optimization.

\textbf{The cost exhibits fine-grained spatial precision.}
In the pushing task, three cubes surround a narrow closed fingertip blade. For the instruction ``push the \{color\} cube,'' candidate trajectories slide the fingertip in a straight line to the endpoint $(x,y)$, requiring contact precise enough to move the named cube by at least 3\,cm. We run 100 CEM trials per instruction across different seeds and execute every selected trajectory. The cost map ranks the target cube first for all three instructions, and \pushTotal\ selected trajectories move it by at least 3\,cm. No unnamed cube moves in any of the 300 trials.

\section{Zero-Shot Planning in Simulation}
\label{sec:droidsim}
\begin{figure*}[t]
    \centering
    \includegraphics[width=\linewidth]{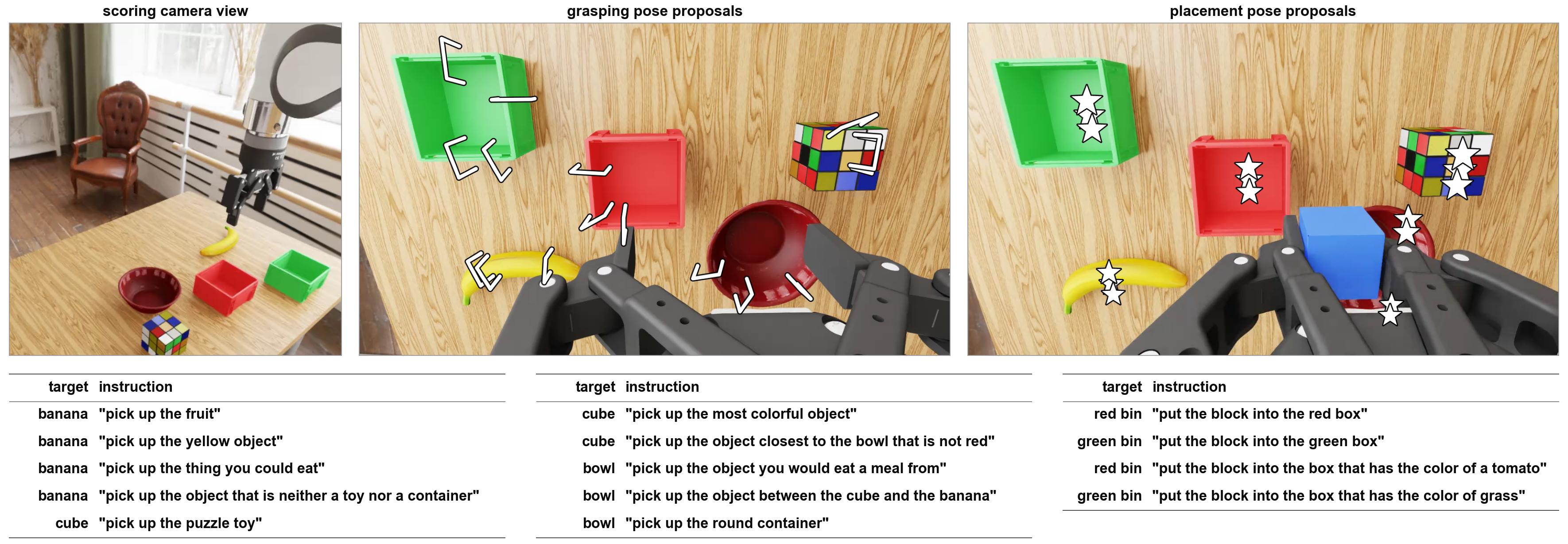}
    \caption{Zero-shot planning in DROID-sim.
    \emph{Left}: The scoring view, containing a banana, a cube, a bowl, and two colored bins.
    \emph{Middle}: 16 language-blind grasp candidates, visualized as gripper outlines from the wrist camera at the home pose.
    \emph{Right}: 16 placement candidates, shown as stars after the block has been grasped.
    \emph{Bottom}: The 14 instructions and the target.
    % The GWM cost provides the sole semantic signal used to rank these geometric candidates for execution.
    }
    \label{fig:droidsim}
    % \vspace{-1em}
\end{figure*}

\textbf{Setup.}
The preceding probes evaluate the cost landscape within a two-dimensional action space.
We now integrate the same GWM and cost formulation into a complete manipulation system, where perception, feasible trajectory generation, and physical execution all play critical roles.
Figure~\ref{fig:droidsim} illustrates the simulated scene and lists the 14 instructions covering 10 picking and 4 placement tasks.
Each task comprises five trials per method, judged by one scripted judge shared by all methods.

\textbf{Methods.} We compare four methods: GWM (ours), TiPToP, $\pi_{0.5}$, and V-JEPA 2-AC.  
The GWM and TiPToP systems share wrist-depth geometry, RANSAC table fitting, M2T2 grasp sampling, and cuRobo motion planning.
\emph{GWM (Ours)}: 
Our proposal mechanism generates candidate actions for every object without reading the instruction, so the GWM cost supplies the sole semantic selection signal.
For picking, the system clusters the wrist depth image into unlabeled objects and samples grasps over the whole scene with M2T2~\citep{yuan2023m2t2}, keeping 16 target poses $T^\ast$ spread across the objects.
cuRobo~\citep{sundaralingam2023curobo} expands each pose into a collision-free joint-space trajectory $a_{t:t+c}=\mathcal{I}(T_t,T^\ast)$, which the GWM cost scores from an external view that matches DROID's shoulder-camera perspective.
Placement poses follow the same geometry-based principle.
The system selects an object by its candidates' aggregate GWM score, then chooses a feasible grasp by M2T2 confidence (Appendix~\ref{app:droidsim}).
Scoring 16 structured candidates per decision keeps open-scene execution within our inference budget.
\emph{TiPToP}~\citep{shen2026tiptop}: 
Uses the proprietary frontier VLM Gemini-Robotics-ER~\citep{geminiroboticsteam2025geminiroboticsbringingai} and open-vocabulary SegmentAnything~\citep{kirillov2023segment} segmentation to produce symbolic goals.
\emph{$\pi_{0.5}$}~\citep{intelligence2025pi05visionlanguageactionmodelopenworld}: 
The official DROID joint-position checkpoint, evaluated zero-shot.
\emph{V-JEPA~2-AC}~\citep{assran2025vjepa2selfsupervisedvideo}: 
A DROID-trained world model from the image-goal lineage.
It receives a pre-captured goal image and scores all 16 trajectories in the same candidate pools.

\begin{wraptable}{r}{0.48\linewidth}
  \vspace{-1.2em}
  \centering
  \footnotesize
  \caption{Success over 14 tasks and five trials per task, with failures attributed where the method exposes the distinction. $\pi_{0.5}$ is scored under a relaxed pick criterion (Appendix~\ref{app:droidsim}).}
  \label{tab:droidsim}
  \setlength{\tabcolsep}{3pt}
  \begin{tabular}{lccc}
    \toprule
    Method & Success & \makecell{Grounding\\failures} & \makecell{Planning\\failures} \\
    \midrule
    GWM (Ours) & \textbf{\simGwm} & 0 & 0 \\
    \makecell[l]{Opposite View} & \simCamOne & 5 & 0 \\
    TiPToP & \simTiptopFrac & 1 & 3 \\
    $\pi_{0.5}$ & \simPiFiveRelaxedFrac & \multicolumn{2}{c}{-} \\
    V-JEPA~2-AC & \simVjepa & 45 & 0 \\
    \bottomrule
  \end{tabular}
  \vspace{-0.8em}
\end{wraptable}
\textbf{Results.}
Table~\ref{tab:droidsim} separates \emph{grounding failures}, where a method selects the wrong object or destination or fails to specify a goal, from \emph{planning failures}, where the grasp slips or no valid plan exists.
Planning with GWM completes \simGwm\ from the scoring view and \simCamOne\ from the opposite view, where all five failures fall on one instruction whose target is small and shadowed from that viewpoint. 
% Because the proposals are language-blind and cover every object, each correct object or destination is selected through the GWM cost, which grounds all 14 instructions correctly.
TiPToP reaches \simTiptopFrac\ (\simTiptop). Its failures comprise three grasp slips and one grounding failure, where the VLM failed to produce the structured goal required by cuTAMP~\citep{shen2024differentiable}, so planning never begins. Despite their shared components, the two stacks differ in grounding, grasp selection, and placement construction, so this is a system-level comparison.
An overview of the two stacks is given in Appendix~\ref{app:stacks}.
$\pi_{0.5}$ reaches \simPiFiveRelaxedFrac\ when picks count if the target ever rises above 0.01\,m during the episode, accommodating lifts that are subsequently dropped; Appendix~\ref{app:droidsim} gives the strict score.
V-JEPA~2-AC completes \simVjepa.

\begin{wrapfigure}{r}{0.28\textwidth}
  \centering
  \vspace{-1em}
  \includegraphics[width=\linewidth]{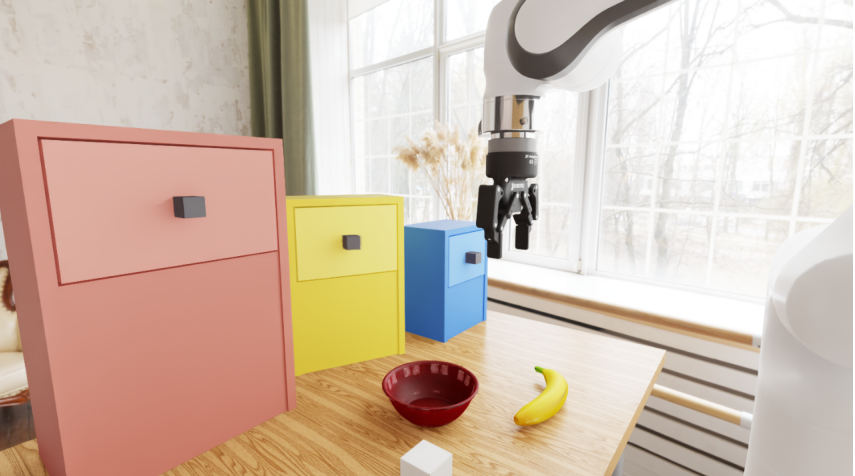}
  % \vspace{-1em}
\end{wrapfigure}
\textbf{Beyond Pick and Place.}
 To test whether the semantic interface itself is tied to only those skills, we score drawer-pull trajectories in a three-cabinet scene. Planning with GWM selects and opens the instructed drawer in \drawerSel\ tasks.
Planning with GWM in new scenes requires only effective trajectory candidates, and no new grounding schema.
TiPToP instead requires new symbolic components to be defined manually for a new skill grounding.
The evaluation protocol is in Appendix~\ref{app:drawer}.

\section{Real-Robot Experiments}
\label{sec:hardware}
\textbf{Setup.}
We deploy the same GWM and planning interface on a physical Franka Panda with a Robotiq 2F-140 gripper, without task-specific adaptation.
% The lab, camera placement, objects, and instructions are absent from training, with no task data, demonstrations, or rewards collected.
% A wrist RGB-D camera supplies geometry to the proposal mechanism, while an external RGB camera supplies the scoring view.
The hardware system retains the simulation stack's components and planning interface. Both systems use the same robot and wrist-mounted RGB-D camera, with scene geometry reconstructed by FoundationStereo~\citep{wen2025foundationstereozeroshotstereomatching}.
Our entire stack runs locally on a single RTX 5090, with GPU memory use within 24\,GB. TiPToP's grounding stack requires a remotely hosted VLM, whose call takes a median of \hwGeminiMedian\,s per decision, while scoring all 16 candidates with GWM takes about 2.5\,s.
Appendix~\ref{app:hardware} gives the full hardware experiment details.

\begin{figure*}[t]
    \centering
    \includegraphics[width=\linewidth]{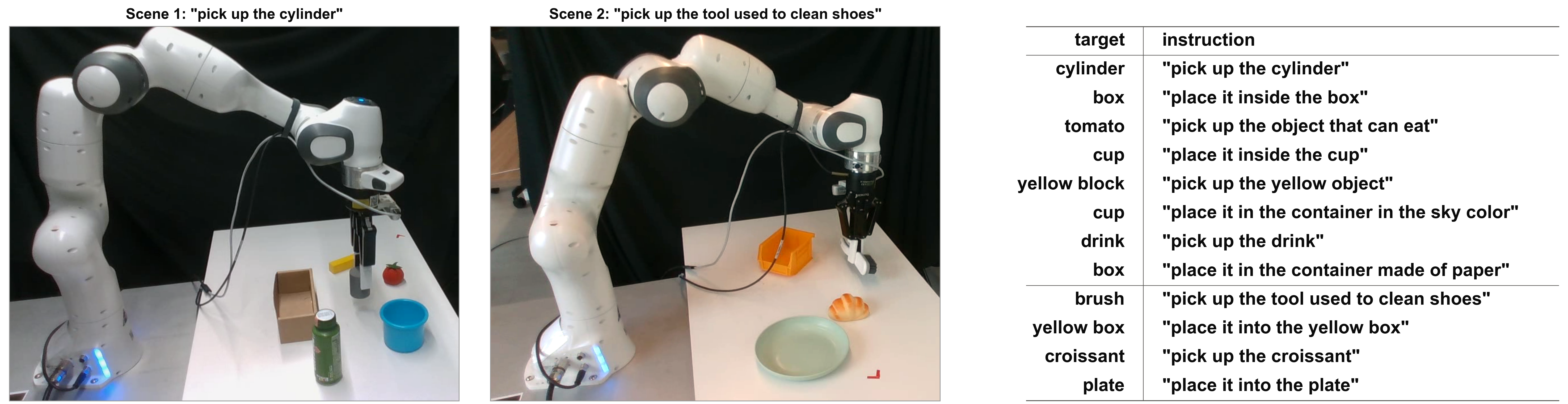}
    \caption{The two real-robot scenes from the external scoring camera, each captured while executing the pick instruction in its title. The figure also lists the twelve instructions and their object or container referents. Scene 1 was recorded under artificial light in the evening, and scene 2 under daylight.}
    \label{fig:hardware}
\end{figure*}

\textbf{Tasks.}
We evaluate pick-and-place in two tabletop scenes, shown with their instructions in Fig.~\ref{fig:hardware}.
% Scene 1 contains four objects, two containers, four pick instructions and four place instructions.
% Scene 2 contains two object-destination pairs, the shoe-cleaning tool into the yellow box and the croissant onto the plate.
The robot plans and executes the pick, re-observes, and then plans and executes the place.
Following TiPToP, trajectories are executed open-loop.
Pick and place sub-tasks are scored separately. Five trials per instruction give 60 trials per method, comprising 40 in scene 1 and 20 in scene 2.
We exclude rounds where a sensor or driver fault prevents planning.

\begin{wraptable}{r}{0.52\linewidth}
  \vspace{-1.2em}
  \centering
  \footnotesize
  \caption{Real-robot success over 60 separately scored pick-and-place sub-task trials per method.}
  \label{tab:hardware}
  \setlength{\tabcolsep}{2pt}
  \begin{tabular}{lccccc}
    \toprule
    Method & Scene 1 & Scene 2 & Total & \makecell{Ground.\\fail.} & \makecell{Plan.\\fail.} \\
    \midrule
    GWM (Ours) & \hwGwmSceneOne & \hwGwmSceneTwo & \textbf{\hwGwmTotal} & 2 & 3 \\
    TiPToP & \hwTiptopSceneOne & \hwTiptopSceneTwo & \hwTiptopTotal & 0 & 8 \\
    \bottomrule
  \end{tabular}
  \vspace{-0.8em}
\end{wraptable}

\textbf{Results.}
Planning with GWM completes \hwGwmTotal\ sub-task trials, with \hwGwmSceneOne\ in scene 1 and \hwGwmSceneTwo\ in scene 2, and TiPToP completes \hwTiptopTotal\ (Table~\ref{tab:hardware}), so the two systems are comparable while only ours runs locally.
TiPToP makes no grounding errors, but all eight failures arise in planning. Three are placement drops in scene 1, where it succeeds on 20/20 picks and 17/20 placements. In scene 2, ``place into the yellow box'' produces no plan in five trials, while every pick and plate placement succeeds.
GWM's five failures comprise three planning and two grounding failures. In both grounding failures, it chooses a distractor close to the target.
Successful instructions require function, material, and indirect color to be resolved against novel objects. Examples include ``the object that can eat,'' ``made of paper,'' and ``the sky color.'' Planning with GWM resolves these references by comparing language with imagined futures through the GWM cost, without a semantic detector, open-vocabulary segmenter, or generative VLM query.
On scene 1, the system completes \hwShoulder\ sub-task trials from an alternative shoulder-height scoring view, compared with \hwGwmSceneOne\ from the reported view (Appendix~\ref{app:cameras}).

\section{Conclusion}
\label{sec:conclusion}
We built GWM from the frozen vision encoder of a video-language embedding model and a learned predictor that approximates state transitions in its visual latent space. 
A fixed pretrained readout maps GWM's predicted futures and language goals into the same embedding space, where negative cosine similarity supplies the planning cost.
This preserves the separation between task-agnostic dynamics and task-specific goals.
The experiments trace this construction from controlled diagnosis to physical deployment.
% On 288 WISER tasks with unseen instructions, predicted futures retain most of the grounding available from real futures, reaching \wiserGwm\ against the Oracle's \wiserOracle. After scaling to real-world data, the GWM cost supports correct semantic grounding and contact-accurate action selection. It then ranks language-blind structured proposals inside complete manipulation systems and matches the modular frontier-VLM system at the system level, \simGwm\ next to its \simTiptopFrac\ in simulation and \hwGwmTotal\ next to \hwTiptopTotal\ separately scored sub-task trials on a real robot in a new lab, while the entire stack runs on a single local GPU without generative VLM queries.
On WISER, planning with predicted futures generalizes to unseen instructions and visual conditions. When trained on real robot data, the same model supports zero-shot manipulation in simulation and on hardware, achieving performance comparable to a frontier-VLM-based planner.

% while replacing a test-time goal image with an instruction.

\textbf{Limitations.} When integrating into a planner, the system depends on a module to propose executable trajectory candidates.
Thus the quality of proposals matters.
The grounding accuracy also depends on the video-language embedding model's ability to recognize robot behavior.
Training a more powerful base model and building GWM in its latent space is a natural next step.
% Our open-scene results cover one simulated layout and two hardware scenes, so broader deployment remains to be tested. 
% Richer proposal distributions, feedback during execution, and longer-horizon language goals are natural next steps.
% Richer proposal distributions and longer-horizon language goals are natural next steps.
% Richer proposal distributions and longer-horizon language goals are natural next steps.

\bibliographystyle{iclr2027_conference}
\bibliography{cite}
\include{appendix}

\end{document}

%% file: appendix.tex
\appendix

\section{Model and Training Details}
\label{app:model}
Throughout the appendix, the \emph{score} of a candidate means the cosine similarity used in Eq.~\ref{eq:cost}, which is the negative of the planning cost.
Candidates are ranked by score and never thresholded.

\textbf{Architecture.}
With Qwen3-VL-Embedding-8B, the predictor $P_\theta$ is a standard transformer operating on the 1{,}620-token latent of a $K{=}6$-frame clip. The WEMM variant uses 336 tokens per clip with the same 4{,}096-dimensional token embeddings.
Its configuration and training hyperparameters are in Table~\ref{tab:hyper-parameter}.
The latent is the token-level internal visual representation of the embedding model, not the pooled vector, so $e_t$, $p_t$, and $\bar{e}_t$ share the same shape and $p_t$ can be fed to the fixed readout $B$ without projection layers.
Fig.~\ref{fig:gwm_cfg_model} contrasts RAT with the learned action-encoder variant ablated in Appendix~\ref{app:wiser}.

\begin{figure*}[ht]
    \centering
    \includegraphics[width=0.95\linewidth]{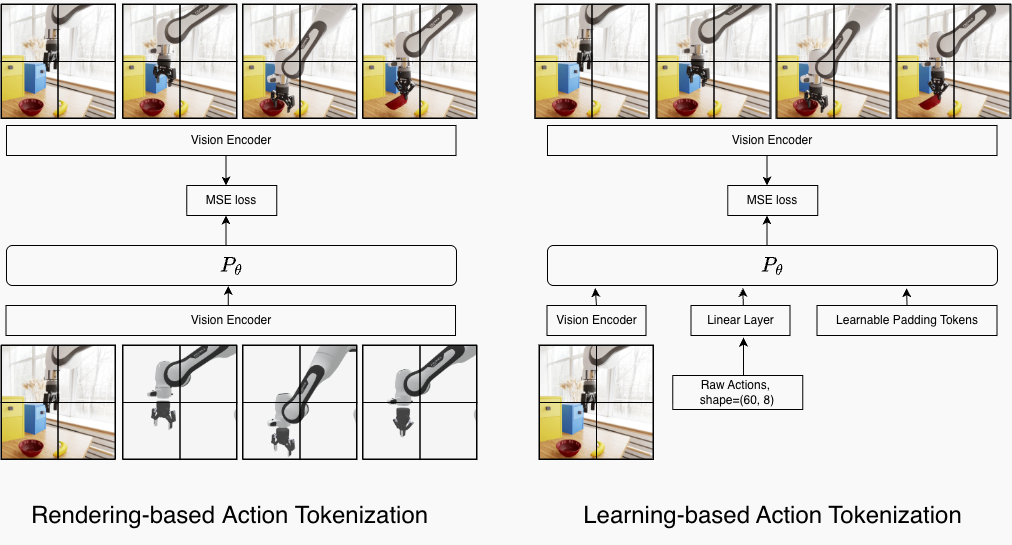}
    \caption{GWM (left) and its raw-action-conditioned variant (right).
    With RAT, actions are rendered into robot-only frames and read by the same frozen vision encoder as the observation. The variant instead encodes numeric actions with a learnable module.
    Captured images are exemplary.}
    \label{fig:gwm_cfg_model}
\end{figure*}

\begin{table}[!ht]
\centering
\caption{Transformer configuration and training hyperparameters of GWM.}
\begin{tabular}{l l}
\toprule
\textbf{Hyperparameter} & \textbf{Value} \\
\midrule
\multicolumn{2}{l}{\textit{Architecture}} \\
Hidden dimension ($d_\text{model}$) & 4096 \\
FFN intermediate dimension ($d_\text{ffn}$) & 8192 \\
Attention head dimension ($d_\text{head}$) & 128 \\
Number of layers & 5 \\
Number of attention heads & 32 \\
Number of KV heads (GQA) & 8 \\
Input / Output dimension & 4096 \\
Input sequence length (Qwen / WEMM) & 1620 / 336 \\
Positional encoding & 2D RoPE \\
Normalization & RMSNorm ($\epsilon = 10^{-5}$) \\
FFN activation & SwiGLU \\
Precision & bfloat16 \\
\midrule
\multicolumn{2}{l}{\textit{Training}} \\
Optimizer & Muon~\citep{jordan2024muon} + Adam~\citep{kingma2017adammethodstochasticoptimization} \\
Learning rate (Muon, hidden weights) & 0.01 \\
Learning rate (Adam, embed/head) & $5 \times 10^{-5}$ \\
Adam $\beta$ & (0.9, 0.95) \\
Weight decay & 0.01 \\
LR scheduler & Cosine annealing \\
Min learning rate & $10^{-6}$ \\
Epochs (WISER model) & 3 \\
Epochs (real-world model) & 10 \\
Gradient clipping & 1.0 \\
Loss function & MSE \\
\bottomrule
\end{tabular}
\label{tab:hyper-parameter}
\end{table}

\textbf{WISER training corpus.}
The model of \S\ref{sec:preliminary} is trained on the 1{,}728 WISER demonstrations, with six per training task.
Control runs at 20\,Hz, and episodes are capped at 120 steps.
Training retains only the main-camera stream, discarding the language labels and wrist camera.
The prediction horizon is $c = 60$ steps, subsampled to $K{=}6$ keyframes.
Training runs for three epochs and takes 20 H100 GPU-hours.

\textbf{Real-world training corpus.}
The model of \S\ref{sec:scaling}-\S\ref{sec:hardware} is trained on two sources with no language labels: (i) the DROID manipulation corpus~\citep{khazatsky2024droid} (74{,}604 episodes re-released at 15\,fps), and (ii) the Franka pick-and-place subset of the MolmoB0T simulation corpus~\citep{deshpande2026molmob0t} ($\sim$35k episodes with shoulder- and side-view streams).
From DROID we keep the external camera streams for which the calibration could be recovered.
Wrist cameras are excluded throughout.
Each training example is a $K{=}6$-frame window spanning roughly three seconds, with log-uniform time-scale augmentation for robustness to different prediction horizons and color jitter on the full-RGB frames ($p = 0.5$).
We hold out 2\% of episodes for validation.
Training runs 34k steps at 384 windows per step, about ten epochs, on 12 H100 GPUs.

\textbf{Rendering.}
The same robot-only renderer produces the RAT frames for training data and inference, and given future joint states the robot mesh is rendered on a black background with the scoring camera's intrinsics and extrinsics.
Only a kinematic model (URDF) and the camera calibration are needed, which is what makes RAT embodiment-agnostic.
Concurrent work applies a related pixel-space action interface to a fine-tuned video generator under the name masked visual actions~\citep{alzayer2026maskedvisualactionsunified}, expressing an action as a partially revealed trajectory in the frame.

\begin{figure}[t]
    \centering
    \includegraphics[width=0.95\linewidth]{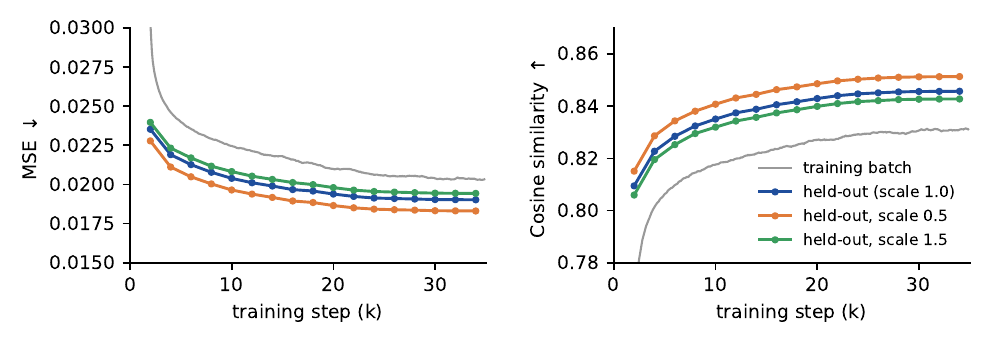}
    \caption{Held-out MSE and cosine similarity between predicted and ground-truth future latents over the real-world training run, evaluated every 2k steps on 256 held-out windows at time scales 1.0, 0.5, and 1.5. The gray curve is the training-batch value smoothed over 200 steps.}
    \label{fig:curves}
\end{figure}
\textbf{Training curves.}
Fig.~\ref{fig:curves} shows the real-world run.
Held-out MSE falls from 0.0235 (step 2k) to 0.0190 (step 34k) and held-out cosine similarity between the predicted and ground-truth future latents rises from 0.809 to 0.846, improving monotonically at all 17 evaluation points, so the prediction steadily approaches the representation of the real future video, which is the property the planning cost relies on (\S\ref{sec:gwm}).
Held-out windows resampled at time scales 0.5 and 1.5 reach cosine 0.851 and 0.843, so the model predicts both faster and slower motion than the nominal window.
Held-out cosine slightly exceeds the training-batch cosine of 0.830, whose batches carry color jitter and random time scales; these different augmentation conditions limit direct comparison of the two values.

\textbf{Goal query.}
To obtain $z_g$, we feed a multimodal prompt into the embedding model.
A retrieval-oriented system prompt $s$ precedes the task instruction $\ell$: \textit{``Retrieve the video which can best finish the manipulation task specified by the user, given the layout of the workspace and the current frame observation.''}
This system prompt steers the model to produce embeddings that align task descriptions with future visual outcomes and discloses no task-specific information.
The main text absorbs $s$ into $\ell$ and writes $z_g = B(\ell, E(o_t))$; this appendix writes $B(s, \ell, E(o_t))$ where the prompt matters.

\textbf{Verbatim instructions in the open-scene experiments.}
In the IsaacSim evaluation (\S\ref{sec:droidsim}) and on the real robot (\S\ref{sec:hardware}), the instruction is used \emph{verbatim}, giving $z_g = B(s, \ell, E(o_t))$ with the current frame from the scoring view and the same system prompt, and none of the sub-task decomposition or softmax normalization the preliminary experiments use (Appendix~\ref{app:prompt}).
On the robot, pick and place are sequential sub-tasks, each receiving one grounding call to score its candidate trajectories.

\section{The WISER Testbed and Full Preliminary Results}
\label{app:wiser}

\subsection{Testbed Description}
WISER (\textbf{W}orld-knowledge \textbf{I}ntegrated \textbf{S}emantic \textbf{E}mbodied \textbf{R}easoning) is described in \S\ref{sec:preliminary} and Fig.~\ref{fig:wiser}.
This section gives the collection protocol, the metrics, and the results held out of the main text.

The testbed is built in ManiSkill~\citep{tao2025maniskill3gpuparallelizedrobotics}.
Demonstrations are collected on training tasks only, with a Franka Panda driven by a privileged motion planner~\citep{mplib} at 20\,Hz (simulation at 100\,Hz).
Six demonstrations per task with randomized initial states give 1{,}728 trajectories.
We record the main camera ($224 \times 448$), wrist camera ($128 \times 128$), joint positions, gripper state, instruction, and actions.
Episodes are capped at 120 steps (6\,s).
The dataset is released in LeRobot~\citep{cadene2026lerobot} and RLDS~\citep{ramos2021rlds} formats.
Task images are AI-generated to avoid copyright issues.
The full task gallery is in Appendix~\ref{app:wiser_gallery}.

\textbf{Metrics.}
\textit{Grasp}: the correct cube is picked.
\textit{Reach}: the tool center point reaches the destination with near-zero speed, even if the grasp failed.
\textit{Success} $=$ \textit{Grasp} $\times$ \textit{Reach}.
Each task is evaluated once and metrics are averaged over the 288 tasks of a split.

\subsection{Planning Hyperparameters}
\begin{figure*}[t]
    \centering
    \includegraphics[width=0.99\linewidth]{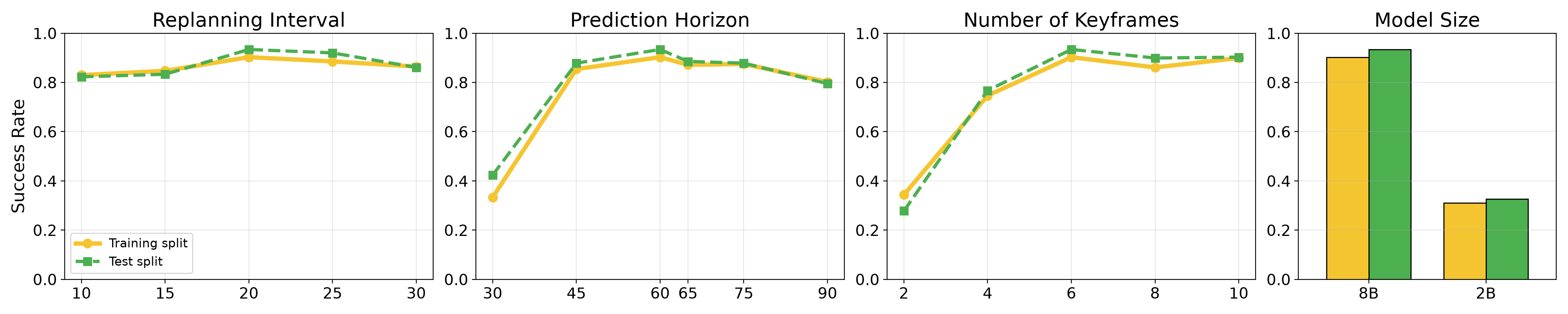}
    \caption{Oracle sweeps of the replanning interval, prediction horizon, and keyframe count on the WISER training and test splits, and the Oracle in the 8B and 2B embedding spaces. The training split selects a replanning interval of 20, $c = 60$, and $K = 6$, and the test split selects the same optimum in each sweep.}
    \label{fig:ablation}
\end{figure*}
The Oracle evaluates whether the fixed readout can rank real future clips under each planning configuration, so it lets us select these hyperparameters independently of the trained predictor.
We select them on the training split and report the test split alongside as a robustness check.
Fig.~\ref{fig:ablation} sweeps the replanning interval over $\{10, 15, 20, 25, 30\}$, the prediction horizon over $\{30, 45, 60, 65, 75, 90\}$, and the keyframe count over $\{2, 4, 6, 8, 10\}$, one at a time with the other two held at their defaults.
The training split selects a replanning interval of 20, $c = 60$, and $K = 6$, and the test split selects the same optimum in each sweep.
On the training split, the optimum sits on a plateau, since an interval of 20 leads 25 by 0.017 and $K = 6$ and $K = 10$ differ by a single episode.
Short horizons and only 2-4 keyframes give lower success, consistent with the need for enough of the future motion to distinguish the intended outcome.

\subsection{Goal Query and Prompt Decomposition}
\label{app:prompt}
On this testbed the instruction $\ell$ follows the template \textit{``Pick up the \{X\} and place it onto the \{Y\}''} and is split by pattern matching into $\ell_{\text{pick}}=$ \textit{``Pick up the \{X\} from the table''} and $\ell_{\text{place}}=$ \textit{``Place the grasped object to the \{Y\} on the table''}.
Each sub-task prompt is embedded with the system prompt, the initial observation $o_0$, and the current observation $o_t$, giving $z_g^{\text{pick}} = B(s, \ell_{\text{pick}}, E(o_0), E(o_t))$ and $z_g^{\text{place}} = B(s, \ell_{\text{place}}, E(o_0), E(o_t))$.
Candidates are scored with Eq.~\ref{eq:cost} against the active sub-goal, and the gripper's contact state selects which sub-goal is active, $z_g^{\text{pick}}$ before the cube is grasped and $z_g^{\text{place}}$ after.

\begin{table}[!ht]
  \small
  \centering
  \caption{Prompt decomposition for GWM (Ours) and InstructVLA on WISER.}
 \vspace{0.5em}
  \label{tab:prompt_decompose}
  \setlength{\tabcolsep}{8pt}
  \begin{tabular}{ccccccc}
    \toprule
    \multirow{2}{*}{Prompt} & \multicolumn{3}{c}{Training Set} & \multicolumn{3}{c}{Test Set} \\
    \cmidrule(lr){2-4} \cmidrule(lr){5-7}
    & Grasp & Reach & Success & Grasp & Reach & Success \\
    \midrule
    GWM $\ell_{\text{pick}}$ + $\ell_{\text{place}}$ & 0.97 & 0.95 & 0.92 & 0.99 & 0.88 & 0.87 \\
    GWM \centering{$\ell$} & 0.93 & 0.88 & 0.82 & 0.92 & 0.76 & 0.73 \\
    \midrule
    InstructVLA $\ell_{\text{pick}}$ + $\ell_{\text{place}}$ & 0.98 & 0.53 & 0.52 & 0.80 & 0.30 & 0.30 \\
    InstructVLA \centering{$\ell$} & 0.98 & 0.92 & 0.89 & 0.79 & 0.51 & 0.47 \\
    \bottomrule
  \end{tabular}
\end{table}

Decomposing the instruction this way raises the test success of planning with GWM from 0.73 to 0.87, supporting the use of atomic sub-task queries for this testbed.
Applying the same decomposition to the best VLA, InstructVLA, \emph{drops} its performance on both splits (train $0.89 \rightarrow 0.52$, test $0.47 \rightarrow 0.30$), showing sensitivity to a change in prompt structure after fine-tuning (Table~\ref{tab:prompt_decompose}).

\subsection{Embedding-Model Ablation}
\label{app:encoder}
\begin{table}[!ht]
  \small
  \centering
  \caption{The same Oracle planner built in different embedding spaces (WISER test split).}
 \vspace{0.5em}
  \label{tab:encoder_ablation}
  \begin{tabular}{lccc}
    \toprule
    Embedding space & Grasp & Reach & Success \\
    \midrule
    Qwen3-VL-Embedding-8B & 1.00 & 0.93 & \textbf{0.93} \\
    WEMM-Embedding-9B~\citep{wemm-embedding} & 0.94 & 0.90 & 0.88 \\
    Qwen3-VL-Embedding-2B & 0.67 & 0.40 & 0.33 \\
    Perception Encoder~\citep{bolya2025perceptionencoderbestvisual} & - & - & 0.00 \\
    \bottomrule
  \end{tabular}
\end{table}
Which spaces can host a grounded world model?
With ground-truth futures supplied by the Oracle, the only variable is the embedding space.
The 8B model reaches \wiserOracle, WEMM-Embedding-9B reaches \wiserWemmOracle, and the 2B sibling drops to \wiserTwoB.
A GWM trained in the WEMM space with the unchanged recipe reaches \wiserWemmTrain\ on the training split and \wiserWemm\ on the test split (Table~\ref{tab:wiser}), demonstrating that the method also works with another video-language embedding model.
We use Qwen3-VL-Embedding-8B in the subsequent experiments because it gives higher WISER test success.
The CLIP-style contrastive Perception Encoder completes \wiserPE\ of the tasks under the same Oracle protocol (Table~\ref{tab:encoder_ablation}).
These comparisons indicate that the embedding space's ability to rank robot futures is an important consideration when choosing a model.

\subsection{The Form and Role of the Predictor}
\label{app:predictor}
Three variants change whether $P_\theta$ is there at all, what space it predicts in, and what it predicts, holding the demonstrations, the twelve candidates, the embedding model, and the cost fixed.

\textbf{No prediction.}
Scoring the condition latent $e_t = E(o_t, R(a_{t:t+c}))$ directly, without mapping it forward, gives 8\% training and \wiserNoPred\ test success, close to the $1/12 = 8.3\%$ rate of picking one of the twelve candidates at random.
The actions are not invisible to the encoder here, since the condition clip contains the rendered trajectory.
What it does not contain is the scene's response.
The rendered frames carry the robot alone, so the cube's new position, which is what the instruction is about, exists only in the predicted future.

\textbf{Prediction in pixels.}
DreamDojo~\citep{gao2026dreamdojogeneralistrobotworld} is an action-conditioned video world model.
We train it on the same demonstrations, generate each candidate's future video, re-encode that video with the same frozen embedding model, and score it with the same cosine, comparing a pixel-prediction pipeline with GWM's latent prediction under the same proposal and scoring protocol.
It reaches 15\% training and \wiserDreamDojo\ test success.
The generated videos are visually plausible but do not follow the commanded actions, so their re-encoded latents barely separate the twelve candidates.
The substitution is also the expensive direction at planning time, since every candidate requires synthesizing a video and running the vision encoder over it, where GWM requires one forward pass of $P_\theta$ on a condition latent the encoder has already produced.

\textbf{Prediction of the pooled embedding.}
The pooled-target variant keeps the 747M predictor and its training recipe and changes only the target and the head.
The backbone output is mean-pooled over the 1{,}620 tokens and mapped by the same linear head to the pooled embedding $B(\bar{e}_t)$ that the frozen readout assigns to the real future clip. Training minimizes MSE against this unit-norm pooled target, and planning scores the predicted vector against $z_g$ directly, so the readout is never applied to a prediction.
It matches GWM on the training split at \wiserZdirectTrain\ and falls to \wiserZdirect\ on the test split, with 0.91 Grasp and 0.66 Reach, compared with GWM's 0.99 and 0.88 (Table~\ref{tab:wiser_full}).
The larger drop in Reach is consistent with the motivation for retaining spatial and temporal structure in the target, although this comparison does not isolate why the pooled target generalizes less well.
The token-level target supervises the layout across the six keyframes, whereas the pooled target compresses the clip into one vector.
Together, the three variants support predicting token-level future latents, which preserves the fixed readout interface and avoids video synthesis during planning.
Table~\ref{tab:wiser_full} gives every variant with all three metrics.
The twelve candidates, the embedding model, and the cosine comparison are shared across these variants.

The no-prediction and pixel-prediction variants also show similar secondary metrics.
Their test \textit{Reach} is 0.44 for both, below GWM's 0.88, and their \textit{Grasp} is 0.26 and 0.28, close to the one-in-four chance of selecting the referred cube.
Executable proposals alone therefore do not ensure that the selected motion satisfies either part of the instruction.

\subsection{Action Tokenization and Cross-Embodiment Generalization}
\begin{wrapfigure}{r}{0.34\textwidth}
  \centering
  \vspace{-1em}
  \includegraphics[width=0.34\textwidth]{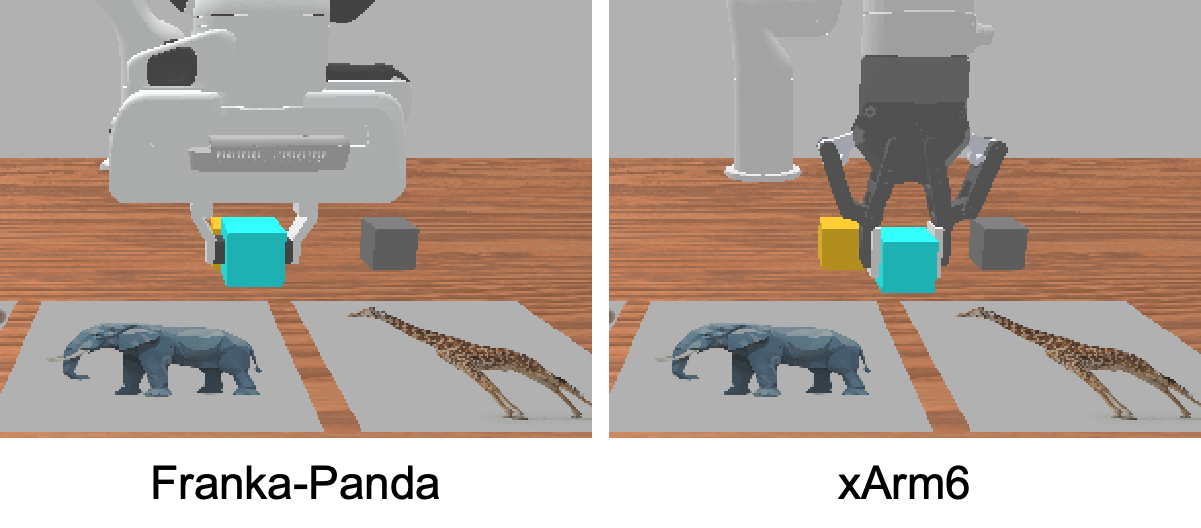}
  \vspace{-2em}
\end{wrapfigure}
The learned-action-encoder variant (Fig.~\ref{fig:gwm_cfg_model}) replaces RAT with a module that embeds raw numeric actions.
It exhibits the same train-test gap as the VLAs (0.74 train, 0.24 test), while RAT's rendered actions align with the observation through the same vision encoder and generalize.
RAT also enables zero-shot cross-embodiment transfer. We collect the 12 unique trajectories on an xArm6 (recording only joint positions and gripper state), render them with the xArm URDF, and reuse the Panda-trained GWM unchanged.
Despite the different action space, kinematics, and appearance, planning succeeds on 87\% of training and 83\% of test tasks (Table~\ref{tab:wiser}).

\subsection{Data Efficiency}
\label{app:data_eff}
The reduced-data variant $\frac{1}{2}\mathcal{D}$ uses 144 tasks from 12 of the 24 categories, with one demonstration per task; the fraction denotes category coverage rather than half of the full corpus's 1{,}728 trajectories.
Planning with this GWM reaches 78\% training and \wiserHalfD\ test success, against \wiserGwmTrain\ and \wiserGwm\ for the full corpus (Table~\ref{tab:wiser_full}).
The loss is concentrated in \textit{Reach} (0.74 test, against 0.88), while test \textit{Grasp} is essentially unchanged (0.98 against 0.99), showing that the reduced corpus mainly affects placement success in this comparison.
This result complements recent evidence on the data efficiency of retrieval-based planners~\citep{Dreczkowski_2025}; GWM learns to predict outcomes while action generation remains with the proposal mechanism.

\subsection{Baseline Implementation Details}
\label{app:impl}
Table~\ref{tab:vla_configs} summarizes the evaluated VLAs.
The replanning interval balances feedback with rollout stability.
Most baselines use 20 steps (1\,s); Motus, GR00T-N1.6, and UniVLA use 40 to limit error accumulation, and InstructVLA uses 16 (Table~\ref{tab:vla_configs}).
Across all models we follow the official fine-tuning setups (full-parameter vs.\ action-head vs.\ LoRA) and train until closed-loop performance plateaus on the training scenes before zero-shot evaluation on the test scenes.
We additionally tested OpenVLA-OFT~\citep{kim2025finetuningvisionlanguageactionmodelsoptimizing}, which overfits a single category (12 tasks) to 100\% but degrades to 24\% training success on the full set, so it is excluded.
For GR00T we report the official GR00T-N1.6, which outperformed the LeRobot GR00T-N1.5 port, and exclude its wrist camera input, which hurt performance.

\begin{table}[htbp]
\centering
\caption{Configuration of the evaluated VLAs.}
\label{tab:vla_configs}
\begin{tabular}{lccccc}
\toprule
\textbf{Model} & \textbf{$c$} & \textbf{Replan Interval} & \textbf{Dataloader} & \textbf{Implementation} & \textbf{Action} \\
\midrule
Motus & 48 & 40 & LeRobot & Official & Absolute \\
XVLA & 40 & 20 & LeRobot & LeRobot & Absolute \\
GR00T-N1.6 & 40 & 40 & LeRobot & Official & Relative \\
InstructVLA & 16 & 16 & RLDS & Official & Absolute \\
SmolVLA & 20 & 20 & LeRobot & LeRobot & Absolute \\
Wall-OSS & 20 & 20 & LeRobot & LeRobot & Absolute \\
$\pi_{0.5}$ & 20 & 20 & LeRobot & LeRobot & Absolute \\
$\pi_{0}$ & 20 & 20 & LeRobot & LeRobot & Absolute \\
InternVLA-A1 & 20 & 20 & LeRobot & Official & Relative \\
UniVLA & 40 & 40 & RLDS & Official & Absolute \\
\bottomrule
\end{tabular}
\end{table}

\subsection{Full WISER Results}
\label{app:wiser_full}
\begin{table}[htbp]
  \centering
  \small
  \caption{Full WISER results for the ten fine-tuned VLAs of Fig.~\ref{fig:wiser_bars} and for the GWM variants and diagnostics, with Grasp and Reach alongside Success.
  Trainable models use the full demonstration corpus except the reduced-data variant, whose category coverage and demonstration count are specified in Appendix~\ref{app:data_eff}. The GWM rows use $N{=}12$ retrieved candidates and the same cosine comparison. They use the same frozen embedding model except for the WEMM variant. GWM (Ours) is repeated from Table~\ref{tab:wiser} for reference.}
  \vspace{0.5em}
  \label{tab:wiser_full}
  \setlength{\tabcolsep}{3.4pt}
  \begin{tabular}{lccccccc}
    \toprule
    \multirow{2}{*}{{Method}} & \multirow{2}{*}[-0.5ex]{{\makecell{H100 GPU\\Hours}}} & \multicolumn{3}{c}{Training Set} & \multicolumn{3}{c}{Test Set} \\
    \cmidrule(lr){3-5} \cmidrule(lr){6-8}
    & & Grasp & Reach & Success & Grasp & Reach & Success \\
    \midrule
    GWM (Ours) & 20 & 0.97 & 0.95 & 0.92 & 0.99 & 0.88 & 0.87 \\
    \midrule
    \multicolumn{8}{c}{\textit{Fine-tuned VLAs}} \\
    \midrule
    InstructVLA~\citep{yang2025instructvlavisionlanguageactioninstructiontuning} & 70 & 0.98 & 0.92 & 0.89 & 0.79 & 0.51 & 0.47 \\
    SmolVLA~\citep{shukor2025smolvlavisionlanguageactionmodelaffordable} & 75 & 0.99 & 1.00 & 0.99 & 0.29 & 0.31 & 0.08 \\
    Wall-OSS~\citep{zhai2025ignitingvlmsembodiedspace} & 80 & 1.00 & 1.00 & 1.00 & 0.68 & 0.50 & 0.40 \\
    GR00T-N1.6~\citep{nvidia2025gr00tn1openfoundation} & 100 & 1.00 & 1.00 & 1.00 & 0.72 & 0.18 & 0.18 \\
    InternVLA-A1~\citep{cai2026internvlaa1unifyingunderstandinggeneration} & 100 & 1.00 & 0.91 & 0.88 & 0.63 & 0.40 & 0.26 \\
    $\pi_{0.5}$~\citep{intelligence2025pi05visionlanguageactionmodelopenworld} & 100 & 1.00 & 0.99 & 0.99 & 0.70 & 0.38 & 0.26 \\
    $\pi_{0}$~\citep{black2026pi0visionlanguageactionflowmodel} & 100 & 1.00 & 1.00 & 1.00 & 0.47 & 0.14 & 0.08 \\
    XVLA~\citep{zheng2025xvlasoftpromptedtransformerscalable} & 100 & 1.00 & 0.88 & 0.88 & 0.44 & 0.17 & 0.17 \\
    UniVLA~\citep{bu2025univlalearningacttaskcentric} & 120 & 0.79 & 0.62 & 0.63 & 0.38 & 0.18 & 0.13 \\
    Motus~\citep{bi2025motusunifiedlatentaction} & 300 & 0.78 & 0.72 & 0.72 & 0.34 & 0.14 & 0.14 \\
    \textit{VLA Average} & - & \textit{0.95} & \textit{0.90} & \textit{0.90} & \textit{0.54} & \textit{0.29} & \textit{0.22} \\
    \midrule
    \multicolumn{8}{c}{\textit{GWM variants and diagnostics}} \\
    \midrule
    GWM w/ action encoder & 20 & 0.91 & 0.77 & 0.74 & 0.47 & 0.42 & 0.24 \\
    GWM w/ pooled target & - & 0.97 & 0.94 & 0.92 & 0.91 & 0.66 & 0.61 \\
    GWM w/ WEMM & - & 0.94 & 0.96 & 0.92 & 0.92 & 0.81 & 0.76 \\
    GWM w/ xArm6 & - & 0.96 & 0.91 & 0.87 & 0.97 & 0.86 & 0.83 \\
    GWM w/ $\frac{1}{2}\mathcal{D}$ & 20 & 0.97 & 0.81 & 0.78 & 0.98 & 0.74 & 0.72 \\
    No prediction & - & 0.27 & 0.41 & 0.08 & 0.26 & 0.44 & 0.09 \\
    DreamDojo (video WM)~\citep{gao2026dreamdojogeneralistrobotworld} & 24 & 0.22 & 0.41 & 0.15 & 0.28 & 0.44 & 0.17 \\
    Oracle & - & 0.97 & 0.92 & 0.90 & 1.00 & 0.93 & 0.93 \\
    \bottomrule
  \end{tabular}
\end{table}
Fig.~\ref{fig:wiser_bars} in \S\ref{sec:preliminary} summarizes the fine-tuned VLAs by their Success, and Table~\ref{tab:wiser_full} gives the full Grasp/Reach/Success triples and the training cost of every model.
All VLAs lose substantial success on the test tasks despite these requiring only demonstrated motions.
Some (SmolVLA, $\pi_0$) reach a near-perfect training success rate yet score below the $1/12 = 8.3\%$ chance rate of random trajectory retrieval at test.
The consistent failure mode is grasping the wrong cube or placing onto a random image.
The two best baselines, Wall-OSS and InstructVLA, are pretrained with auxiliary embodied-VQA tasks that help retain VLM knowledge.
Even so, Appendix~\ref{app:eagle} shows InstructVLA's base VLM localizes the correct destination in 81\% of test scenarios before fine-tuning, while the fine-tuned policy reaches it in 51\%.
All VLAs show partial generalization at the cube-picking stage (average test Grasp 0.54), which requires recognizing unseen colors and referring expressions but no world knowledge.
GR00T-N1.6 and InternVLA-A1 act in a relative action space and the other baselines in an absolute one, and the train-test gap appears in both groups.
The baselines cover several VLA recipes (latent-action pretraining, knowledge insulation, mixture-of-transformers, VQA auxiliary tasks, video-action joint training), so the drop is observed across architectures.
The training performance of planning with GWM (\wiserGwmTrain) sits below the best VLAs and close to the Oracle's \wiserOracleTrain.
The Oracle provides a reference for grounding with real future clips, rather than a strict upper bound; predicted futures score slightly higher on training tasks and lower on test tasks (\wiserGwm\ versus \wiserOracle).

\subsection{Visual Grounding of the Base VLM}
\label{app:eagle}
To rule out that WISER's visuals are simply unrecognizable to pretrained VLMs, we probe the base model of the best VLA before fine-tuning.
For each of the 24 test configurations we capture the initial observation and ask Eagle-2B~\citep{li2025eagle2buildingposttraining} which of the three destination images (left, middle, right) best matches the place referring expression.
The base VLM localizes the correct destination in 81\% of the 288 test scenarios, while after fine-tuning on OXE and the WISER training data, InstructVLA's reaching success is 51\%.
This shows that many WISER references are recognizable to the base VLM, while comparing localization with executed reaching does not isolate the effect of fine-tuning.

\section{Real-Data Probe Protocols}
\label{app:cem}

\textbf{Shared CEM setup.}
The pointing and pushing probes in \S\ref{sec:scaling} use separate IsaacSim scenes and the same real-data GWM checkpoint as the subsequent system evaluations.
A two-dimensional sample specifies the $(x,y)$ endpoint of a target pose $T^\ast$, with a fixed height $z_0$ and downward orientation.
The interpolator $\mathcal{I}$ expands each endpoint into a 31-waypoint trajectory from the home pose.
Both probes refine 30 CEM samples over five iterations.
CEM refines the endpoints using the scores of their predicted futures, and Fig.~\ref{fig:scaling} visualizes the scoring landscapes.

\subsection{Pointing}
\label{app:pointing}
Four printed photos depict a dog, a panda, a banana, and a strawberry, and the instruction is ``point at the image of the \{name\}.''
We report two distinct checks, whether the spatial map's lowest-cost position lies on the named photo (\cemGrid) and whether CEM's best sample reaches it (\cemBest).
Fig.~\ref{fig:scaling} shows the cost maps and the paths of the Gaussian means.

\subsection{Pushing}
\label{app:pushing}
Three 4.7\,cm cubes surround a gripper with a 2.4\,cm-wide closed fingertip blade.
For ``push the \{color\} cube,'' the candidate slides the fingertip in a straight line to its sampled endpoint.
We run 100 CEM trials per instruction with different seeds and execute each selected trajectory.
A push succeeds when the named cube moves at least 3\,cm, giving \pushTotal\ successful pushes.
All \pushDir\ successful pushes lie within $45^\circ$ of the ideal sliding direction, with median angular errors of $3.6^\circ$, $4.4^\circ$, and $5.3^\circ$ for the three instructions.

\section{IsaacSim Protocol}
\label{app:droidsim}

\textbf{Scene and tasks.}
The evaluation uses one IsaacSim tabletop scene with a Franka arm, a banana, a cube, and a bowl for the pick tasks, plus a red and a green bin for the place tasks.
Each task has five trials with a 90\,s episode cap.
Per-task results for every arm are in Table~\ref{tab:droidsim_pertask}.

\begin{table}[ht]
  \centering
  \small
  \caption{Per-task results in the IsaacSim evaluation (successes out of 5 trials).
  Cam2 is the shipped scoring view used in \S\ref{sec:droidsim}. Cam1 is the symmetric opposite view. Fusion averages the two views' scores per candidate (Appendix~\ref{app:cameras}).
  The $\pi_{0.5}$ column uses the relaxed pick criterion of this appendix.}
  \vspace{0.4em}
  \label{tab:droidsim_pertask}
  \setlength{\tabcolsep}{3pt}
  \begin{tabular}{llcccccc}
    \toprule
    Task (referring expression) & Target & \makecell{GWM\\cam2} & \makecell{GWM\\cam1} & \makecell{GWM\\fusion} & TiPToP & $\pi_{0.5}$ & \makecell{V-JEPA\\2-AC} \\
    \midrule
    ``pick up the fruit'' & banana & 5/5 & 5/5 & 5/5 & 4/5 & 0/5 & 0/5 \\
    ``\ldots the yellow object'' & banana & 5/5 & 0/5 & 5/5 & 5/5 & 5/5 & 0/5 \\
    ``\ldots the thing you could eat'' & banana & 5/5 & 5/5 & 5/5 & 5/5 & 1/5 & 0/5 \\
    ``\ldots neither a toy nor a container'' & banana & 5/5 & 5/5 & 5/5 & 5/5 & 2/5 & 0/5 \\
    ``\ldots the puzzle toy'' & cube & 5/5 & 5/5 & 5/5 & 4/5 & 5/5 & 0/5 \\
    ``\ldots the most colorful object'' & cube & 5/5 & 5/5 & 5/5 & 4/5 & 5/5 & 0/5 \\
    ``\ldots closest to the bowl that is not red'' & cube & 5/5 & 5/5 & 5/5 & 4/5 & 4/5 & 0/5 \\
    ``\ldots you would eat a meal from'' & bowl & 5/5 & 5/5 & 5/5 & 5/5 & 3/5 & 5/5 \\
    ``\ldots between the cube and the banana'' & bowl & 5/5 & 5/5 & 5/5 & 5/5 & 0/5 & 5/5 \\
    ``\ldots the round container'' & bowl & 5/5 & 5/5 & 5/5 & 5/5 & 5/5 & 5/5 \\
    \midrule
    ``put the block into the red box'' & red bin & 5/5 & 5/5 & 5/5 & 5/5 & 4/5 & 0/5 \\
    ``\ldots the green box'' & green bin & 5/5 & 5/5 & 5/5 & 5/5 & 3/5 & 5/5 \\
    ``\ldots the box that has the color of a tomato'' & red bin & 5/5 & 5/5 & 5/5 & 5/5 & 0/5 & 0/5 \\
    ``\ldots the box that has the color of grass'' & green bin & 5/5 & 5/5 & 5/5 & 5/5 & 0/5 & 5/5 \\
    \midrule
    Total & & \textbf{70/70} & 65/70 & \textbf{70/70} & 66/70 & \simPiFiveRelaxedFrac & 25/70 \\
    \bottomrule
  \end{tabular}
\end{table}

\textbf{Judge.}
The same scripted judge evaluates every arm and was never retuned; the additional relaxed pick criterion reported for $\pi_{0.5}$ is defined below.
Pick: the target's mesh center is at least 0.15\,m above its settled height at episode end.
Place: the block starts held in the gripper and succeeds when its mesh center is within 0.05\,m (xy) of the named bin's center, with its height inside the bin's band, judged from the last pre-reset snapshot.

\textbf{Proposal pipeline.}
Both systems lift wrist depth to a point cloud and fit the table plane with RANSAC.
The GWM system then groups points above the surface into anonymous objects with DBSCAN; TiPToP's object segmentation is described in Appendix~\ref{app:stacks}.
A grasp-synthesis network~\citep{yuan2023m2t2} proposes grasps over the whole scene, and confidence-weighted sampling keeps 16 of them spread across the objects, so every object receives candidates regardless of the instruction.
cuRobo~\citep{sundaralingam2023curobo} expands each grasp into a collision-free joint-space trajectory.
A candidate is a sequence of joint waypoints plus gripper events.

\textbf{Selection.}
Stage 1 picks the object whose candidates have the best aggregate score (mean over the object's candidates in simulation).
Stage 2 picks the grasp within that object by the grasp network's confidence, subject to a feasibility gate.
Placement candidates are built in the same way from geometry-based target poses on each perceived destination (Appendix~\ref{app:stacks}).

\textbf{Fairness and caveats.}
The language-goal methods receive byte-identical instructions and scenes, and the goal-image baseline uses the corresponding scenes and supplied goal images.
$\pi_{0.5}$ is the official DROID joint-position checkpoint, run zero-shot and closed-loop with no simulation fine-tuning.
The comparison is on task outcome, not compute.
This evaluation covers one scene, one layout, and 14 instructions at 5 trials each.
We read the margins as ``grounding is reliable in this scene'', not as transferable percentages.
The scoring view was chosen after inspecting both cameras (Appendix~\ref{app:cameras}).

\textbf{$\pi_{0.5}$ under relaxed criteria.}
Under the official judge $\pi_{0.5}$ scores \simPiFive\ (pick 0/50, place 7/20).
Per-step diagnostics show $\pi_{0.5}$'s pick failures are grasp \emph{retention}, not grounding, since it reaches for the correct object and produces real lifts that are then dropped, ending near table height.
Under progressively relaxed pick criteria, end-state height above 0.10/0.05/0.01\,m gives 0/2/5 of 50, and ``ever lifted above 0.01\,m during the episode'' gives 30 of 50 trials (cube tasks 14/15).
Combining the most relaxed pick criterion with the official place judge yields \simPiFiveRelaxedFrac, \simPiFiveRelaxed, the rate reported in \S\ref{sec:droidsim}.
Its place failures on indirect color references (``the color of a tomato / grass'', 0/10) carry the block toward a bin but drop it on the table, while direct color references mostly succeed (7/10).

\textbf{Goal-image world-model baseline (V-JEPA~2-AC).}
V-JEPA~2-AC~\citep{assran2025vjepa2selfsupervisedvideo} plans toward a goal image with numeric end-effector actions.
We supply both and score all trajectories in the same 16-candidate pools, with the same judges and two-stage selection rule.
Its evaluation differs from the language-goal systems' in the following ways, stated explicitly.
\emph{Goals}: each task receives one goal image, the final scoring-camera frame of the target object's highest-confidence successful trajectory. All 16 candidate trajectories remain available for selection.
\emph{Actions}: each candidate's joint-waypoint plan is converted by forward kinematics into end-effector states at the model's training cadence (one action per 0.267\,s). The resulting per-step deltas lie within the model's action cap.
\emph{Selection}: the candidate's actions are rolled out autoregressively in latent space from the current frame, the energy is the L1 distance between the predicted final frame's features and the goal image's features, and the object is chosen by the same two-stage rule as planning with GWM. The plain per-candidate argmin instead gives 10/70.
The pick and place judges are unchanged from the main protocol.

\subsection{The GWM and TiPToP Stacks Compared}
\label{app:stacks}
Both stacks use wrist-depth geometry, RANSAC table fitting, DBSCAN point-cloud cleanup, convex hulls, M2T2 grasp sampling~\citep{yuan2023m2t2}, and cuRobo motion planning~\citep{sundaralingam2023curobo}. These are shared components. The differences lie in how objects are identified, how grasps are selected, and how placement targets and plans are constructed. The comparison in \S\ref{sec:droidsim} and \S\ref{sec:hardware} therefore evaluates complete systems.

\textbf{Perception.}
Both stacks fit the table plane with RANSAC and use DBSCAN to clean its point cloud. TiPToP obtains object regions from VLM bounding boxes and segmentation masks. The GWM stack instead uses DBSCAN to group above-table points into anonymous objects. Both construct convex hulls for object geometry; the distinction is whether object identification requires semantic labels.

\textbf{Grasp choice.}
Within the object that the GWM cost selects, the GWM stack executes the grasp with the highest M2T2 confidence, after a geometric feasibility check at the closing configuration that rejects grasps whose finger pads would meet too thin a slab or land off the object's center.
The GWM cost selects the object by semantic agreement, and this second stage delegates grasp robustness to the grasp network.
TiPToP selects a feasible grasp as part of its cuTAMP symbolic task plan, rather than applying this confidence-based selection after choosing an object.

\textbf{Placement target.}
The GWM stack measures the object in hand from the point cloud, reads the landing surface of every perceived cluster from the raw points, and tells a hollow container from a solid block by the depth of its interior below the rim and by the rim enclosing that interior all around.
A hollow destination receives candidates on its inner floor.
Each candidate is a two-segment cuRobo plan, an approach above the rim followed by a constrained straight descent, with the destination disabled as an obstacle during the descent only and the closed-finger overhang subtracted from the clearance.
Multiple offsets per destination are planned independently, and each successful plan becomes a candidate.
TiPToP represents the destination by the convex hull of its mask, so an open box becomes a solid block whose inside cannot be told from its top, a limitation its documentation states.
Placement is then a stability constraint on the shrunken bounding box of that hull, solved by cuTAMP with the destination kept as a collision obstacle throughout.
A small open container can therefore admit no placement pose for TiPToP while the GWM stack lands inside it, which is consistent with the five no-plan outcomes on ``place into the yellow box'' in \S\ref{sec:hardware}.

\textbf{Execution.}
In simulation a GWM plan ends holding the object inside the destination, whereas a TiPToP plan releases it and returns home, and the judge accepts both end states.
On hardware both systems release the object and return to the home pose.

% \Needspace{6\baselineskip}
\subsection{Drawer Selection}
\label{app:drawer}
Three single-drawer cabinets, a large muted-red one, a medium yellow one yawed $25^\circ$, and a small blue one, stand behind a block, a bowl, and a banana on the table.
Six candidate trajectories are built with the same inverse-kinematics machinery, three drawer pulls, each comprising an approach, a knob grasp, and a pull along the slide axis, and three top-down grasps of the table objects that serve as distractors.
Each drawer is referred to by two instructions of the form ``open the drawer of the \{attribute\} cabinet'', one naming its color (red, yellow, blue) and the other a size or position attribute (largest, middle, smallest).
Each instruction is averaged over five phrasings.
Planning with GWM selects the instructed drawer pull in \drawerSel\ tasks, and replaying the selected pull in the simulator opens that drawer in each of them.
The miss is ``the smallest cabinet'', where the pull of the middle cabinet scores $0.022$ above the pull of the instructed one; margins here are differences in cosine score between two candidate trajectories under the same instruction.

\section{Real-Robot Protocol}
\label{app:hardware}

\textbf{Rig.}
The rig uses a Franka Panda with a Robotiq 2F-140 gripper.
A wrist-mounted RGB-D camera supplies the images from which FoundationStereo~\citep{wen2025foundationstereozeroshotstereomatching} reconstructs the scene geometry, and one external RGB camera provides the scoring view with its frame cropped to the workspace.
The gripper is a Robotiq 2F-140, whereas most DROID episodes use the Robotiq 2F-85; the renderer simply uses the rig's own URDF, and no adaptation was needed.

\textbf{Pipeline.}
The GWM pipeline follows Appendix~\ref{app:droidsim} using the real wrist point cloud.
The pipeline splits 16 candidates evenly across the discovered objects.
Selection is two-stage with the object aggregate taken as the mean of its top-2 candidate scores, followed by the grasp-confidence choice and a feasibility gate.
Because each sub-task receives one grounding call, the robot plans and executes the pick open-loop, re-observes, then plans and executes the place.
The gripper state routes which sub-task is active.
Trajectories are replayed at reduced speed with an acceleration cap.

\textbf{Scoring efficiency.}
Scoring 16 candidates takes about 4\,s sequentially and 2.5\,s in parallel.
TiPToP's remote VLM call takes a median of \hwGeminiMedian\,s per decision and ranges from \hwGeminiMin\ to \hwGeminiMax\,s over the 20 recorded scene-1 runs.
The whole system (embedding model, predictor, perception, planner) runs on a single RTX 5090, with GPU memory use within 24\,GB.

\textbf{Success criteria and anomaly rule.}
Each pick or place sub-task is labeled separately by a human operator.
The denominator of 60 is the total number of evaluated sub-tasks, comprising 30 picks and 30 placements across the two scenes, with five trials per instruction.
A pick succeeds if the named object is grasped and held.
A place succeeds if the carried object ends inside or on the named destination.
Rounds in which the system never planned due to a sensor or driver fault are excluded for both methods.
Every failure is attributed to \emph{grounding} (wrong object or destination selected) or \emph{planning} (no valid proposal, drop or unstable grasp during execution, no plan produced).

\begin{table}[ht]
  \centering
  \small
  \caption{Per-task real-robot results. Each row reports successes out of 5 trials of one pick or place sub-task; totals count the two stages separately.}
  \vspace{0.4em}
  \label{tab:hw_pertask}
  \setlength{\tabcolsep}{5pt}
  \begin{tabular}{llccc}
    \toprule
    Scene & Task & \makecell{GWM\\best view} & \makecell{GWM\\shoulder view} & TiPToP \\
    \midrule
    1 & pick the cylinder & 5/5 & 5/5 & 5/5 \\
    1 & place it inside the box & 4/5 & 5/5 & 5/5 \\
    1 & pick the object that can eat & 5/5 & 5/5 & 5/5 \\
    1 & place it inside the cup & 4/5 & 3/5 & 5/5 \\
    1 & pick the yellow object & 5/5 & 4/5 & 5/5 \\
    1 & place it in the container in the sky color & 5/5 & 3/5 & 3/5 \\
    1 & pick the drink & 4/5 & 5/5 & 5/5 \\
    1 & place it in the container made of paper & 4/5 & 4/5 & 4/5 \\
    \midrule
    2 & pick the shoe-cleaning tool & 5/5 & - & 5/5 \\
    2 & place it into the yellow box & 4/5 & - & 0/5 \\
    2 & pick the croissant & 5/5 & - & 5/5 \\
    2 & place it onto the plate & 5/5 & - & 5/5 \\
    \midrule
    & Total & \hwGwmTotal & \hwShoulder\ (scene 1) & \hwTiptopTotal \\
    \bottomrule
  \end{tabular}
\end{table}

\textbf{Results.}
Table~\ref{tab:hw_pertask} lists every task.
In scene 1 TiPToP scores 20/20 on the pick sub-tasks and 17/20 on the place sub-tasks.
In scene 2, TiPToP's picks succeed while ``place into the yellow box'' produced no plan in all five trials (planning failures).
The shoulder-view column is the camera-placement comparison of Appendix~\ref{app:cameras}, with the same system, same tasks, and original camera position (two excluded anomaly rounds).

\section{Scoring View and Multi-View Fusion}
\label{app:cameras}
The main-text system comparisons score each candidate from one external camera at a time, using cam2 in IsaacSim and the reported view on the robot. This appendix compares individual scoring views and reports a two-view fusion experiment in IsaacSim.
The scoring view is a first-order factor. RAT renders the candidate with the scoring camera's calibration, so what the embedding model can read depends on what the camera can see.
We evaluate whether averaging scores across views can recover errors caused by limited visibility from one camera.

\textbf{Simulation.}
From the shipped view (cam2), planning with GWM selects correctly on all 14 tasks (\simGwm).
From the symmetric opposite view (cam1), the same system scores \simCamOne, and the five failures are all on ``the yellow object'', where the banana appears small, distant, and shadowed by the gripper from that viewpoint.
The selection error repeats deterministically across trials and is consistent with reduced target visibility from this view.
Averaging the two views' scores per candidate corrects this error and yields \simGwm\ (Table~\ref{tab:droidsim_pertask}, fusion column).

\begin{figure}[t]
  \centering
  \begin{minipage}[t]{0.48\linewidth}
    \centering
    \includegraphics[width=\linewidth]{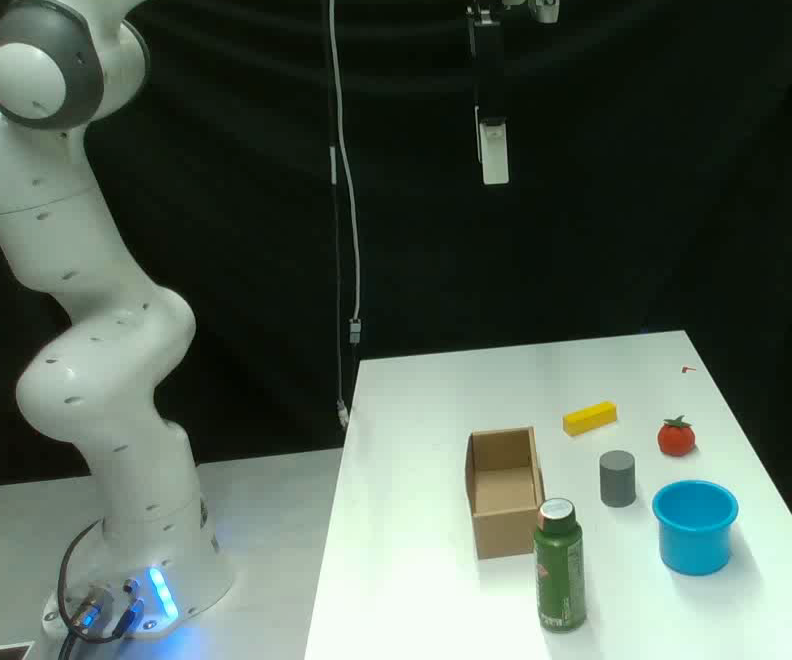}
    \par\smallskip
    \small\textbf{(a) Reported view}\\
    \hwGwmSceneOne\ successful sub-task trials
  \end{minipage}\hfill
  \begin{minipage}[t]{0.48\linewidth}
    \centering
    \includegraphics[width=\linewidth]{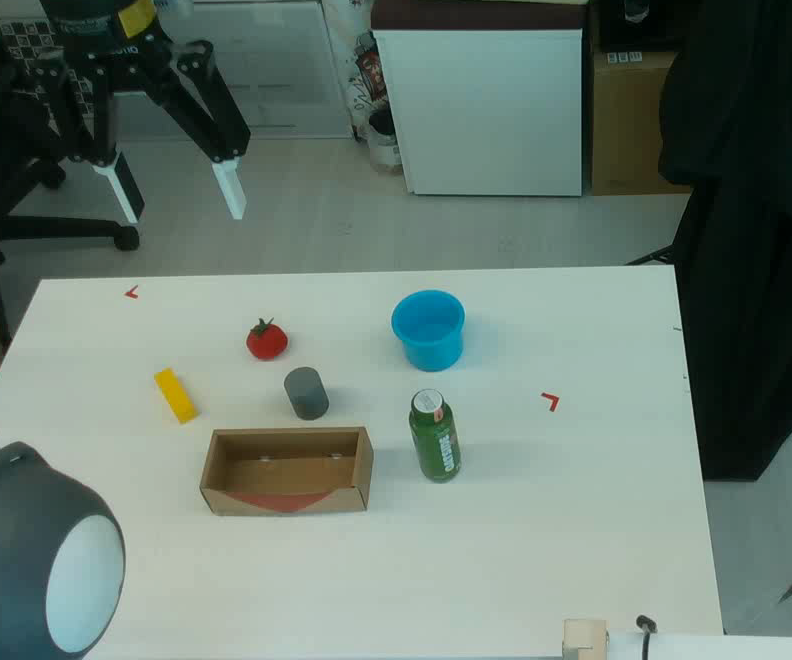}
    \par\smallskip
    \small\textbf{(b) Shoulder-height view}\\
    \hwShoulder\ successful sub-task trials
  \end{minipage}
  \caption{Two scoring views of hardware scene 1, shown at the start of separate ``pick up the cylinder'' trials. Success counts cover all eight scene-1 pick and place instructions, with five trials each. Each evaluation uses one camera.}
  \label{fig:hardware_views}
\end{figure}

\textbf{Real robot.}
Figure~\ref{fig:hardware_views} shows the two scoring views used for the scene-1 hardware experiments.
The same system completes \hwGwmSceneOne\ sub-task trials from the reported view and \hwShoulder\ from the shoulder-height view, with the same task set evaluated separately from each view.
Table~\ref{tab:hw_pertask} gives the per-task results.

\begin{table}[t]
  \centering
  \small
  \caption{Object-selection margins per scoring view. The margin is the aggregate cosine of the best object minus the second best, and \ding{55} marks a wrong selection.}
  \vspace{0.4em}
  \label{tab:margins}
  \setlength{\tabcolsep}{8pt}
  \begin{tabular}{lccc}
    \toprule
    Task & cam2 & cam1 & fusion \\
    \midrule
    fruit & $+0.104$ & $+0.014$ & $+0.057$ \\
    yellow & $+0.105$ & $+0.002$\,\ding{55} & $+0.053$ \\
    eat & $+0.091$ & $+0.018$ & $+0.056$ \\
    negation & $+0.094$ & $+0.019$ & $+0.065$ \\
    puzzle & $+0.052$ & $+0.076$ & $+0.069$ \\
    colorful & $+0.054$ & $+0.057$ & $+0.062$ \\
    nearbowl & $+0.023$ & $+0.016$ & $+0.029$ \\
    eatfrom & $+0.028$ & $+0.029$ & $+0.031$ \\
    between & $+0.015$ & $+0.005$ & $+0.015$ \\
    container & $+0.018$ & $+0.050$ & $+0.042$ \\
    place\_red & $+0.020$ & $+0.039$ & $+0.028$ \\
    place\_green & $+0.084$ & $+0.054$ & $+0.072$ \\
    place\_tomato & $+0.007$ & $+0.029$ & $+0.016$ \\
    place\_grass & $+0.063$ & $+0.041$ & $+0.057$ \\
    \bottomrule
  \end{tabular}
\end{table}

\textbf{Multi-view fusion.}
Because RAT renders per view, the same candidate can be scored from several cameras and the scores averaged, with no retraining and no new parameters.
The two-view fusion renders each candidate under both camera calibrations, scores each rendering against the instruction, and averages the two scores before object aggregation.
Table~\ref{tab:margins} reports the object-selection margin, the aggregate score of the selected object minus the runner-up, for each view and their fusion.
Because fusion precedes object aggregation, its object margin need not equal the mean of the single-view margins.
On ``yellow'', cam1 selects the wrong object with a margin of $+0.002$, whereas fusion selects the correct object with a margin of $+0.053$.
The two individual views select correctly on 13/14 and 14/14 tasks, while fusion selects correctly on all 14, with a minimum object-selection margin of $+0.015$.
Fusion recovers the weaker view's error while preserving all correct selections from the stronger view. It does not consistently increase the margin over the better individual view.
These results demonstrate multi-view fusion as a way to improve robustness to the choice of scoring camera.

\clearpage
\section{WISER Task Gallery}
\label{app:wiser_gallery}
All training and test tasks of WISER are shown below.
The images are AI-generated to avoid copyright issues.
\begin{figure*}[h]
    \centering
    \includegraphics[width=\linewidth]{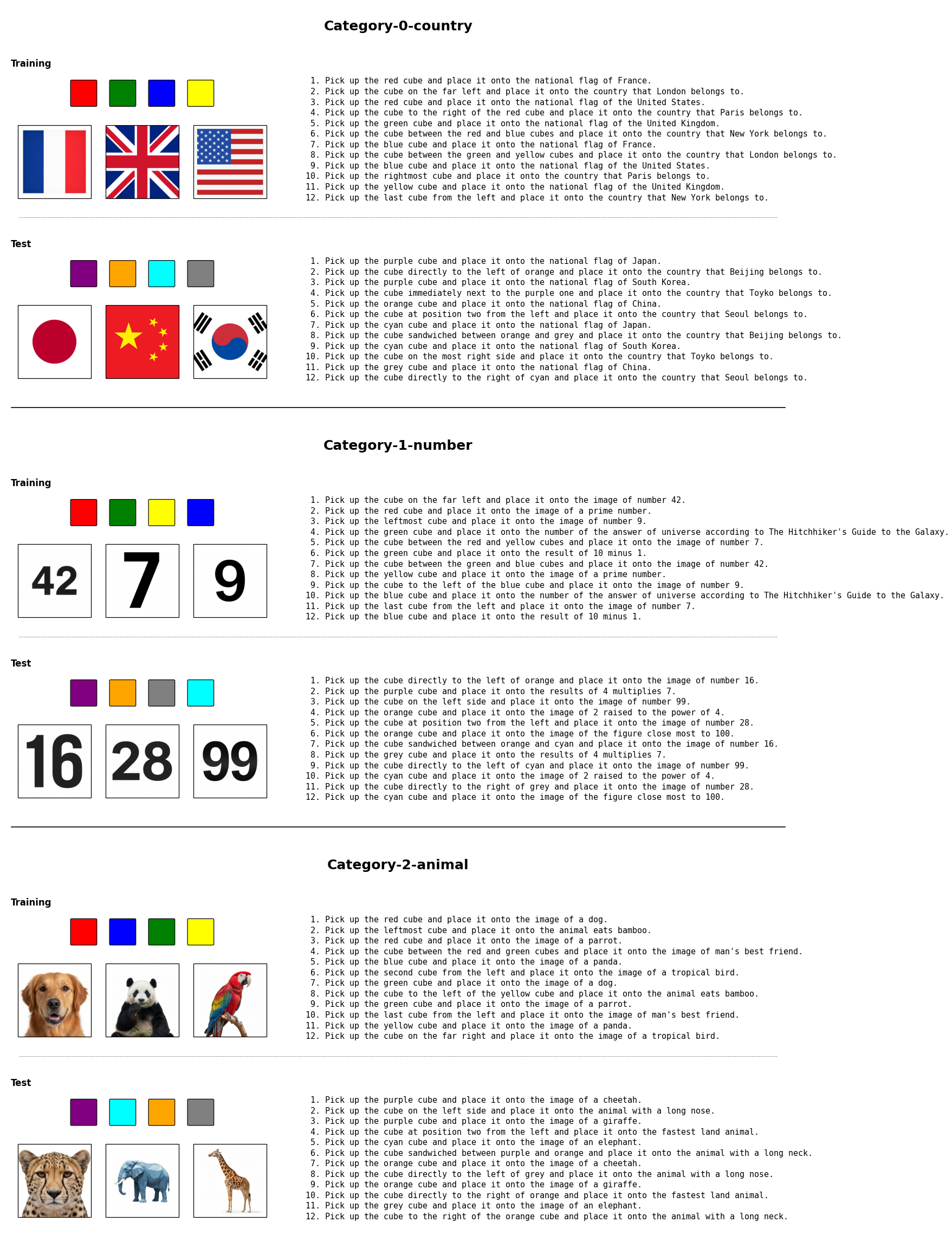}
\end{figure*}
\begin{figure*}
    \centering
    \includegraphics[width=\linewidth]{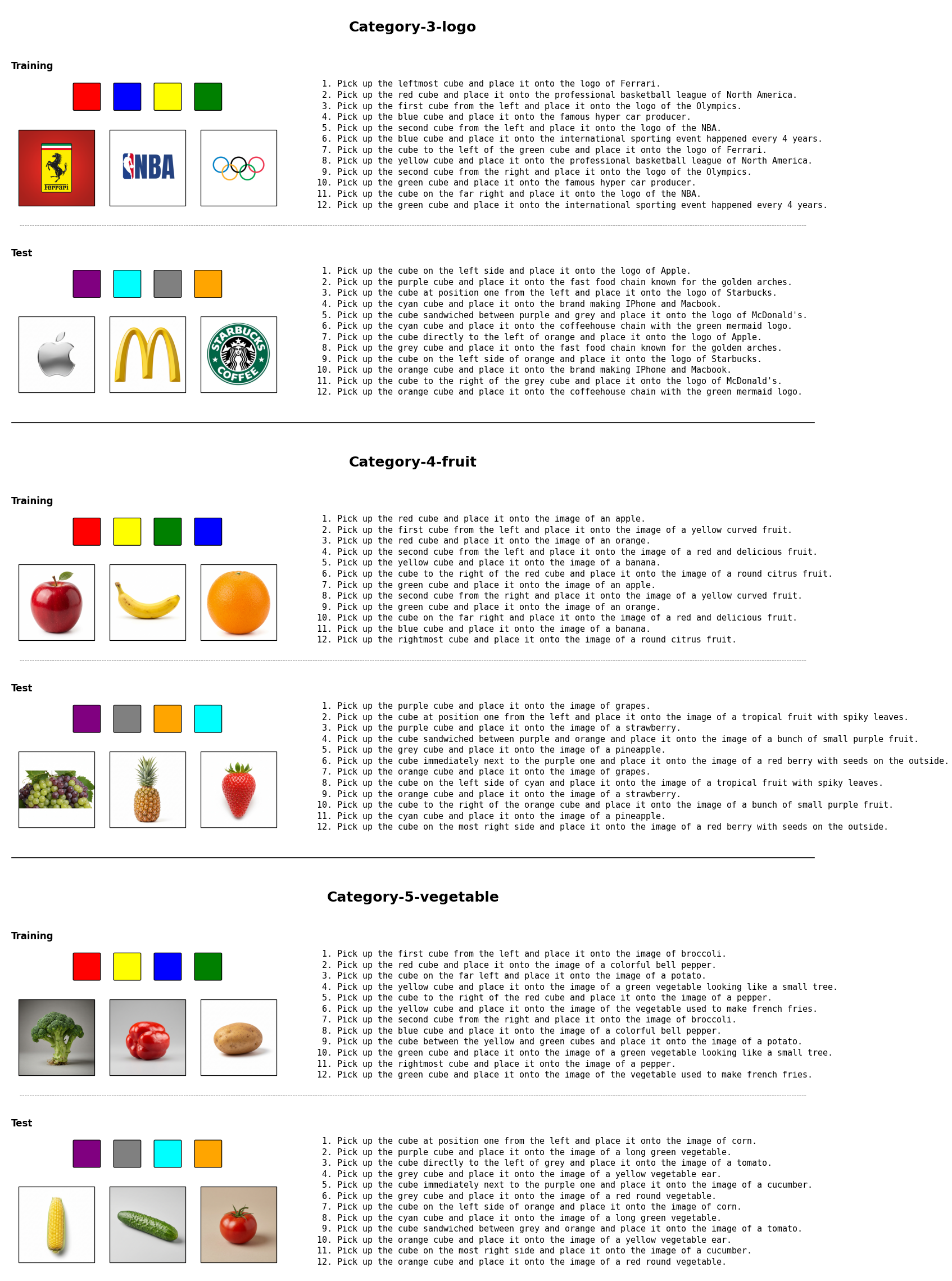}
\end{figure*}
\begin{figure*}
    \centering
    \includegraphics[width=\linewidth]{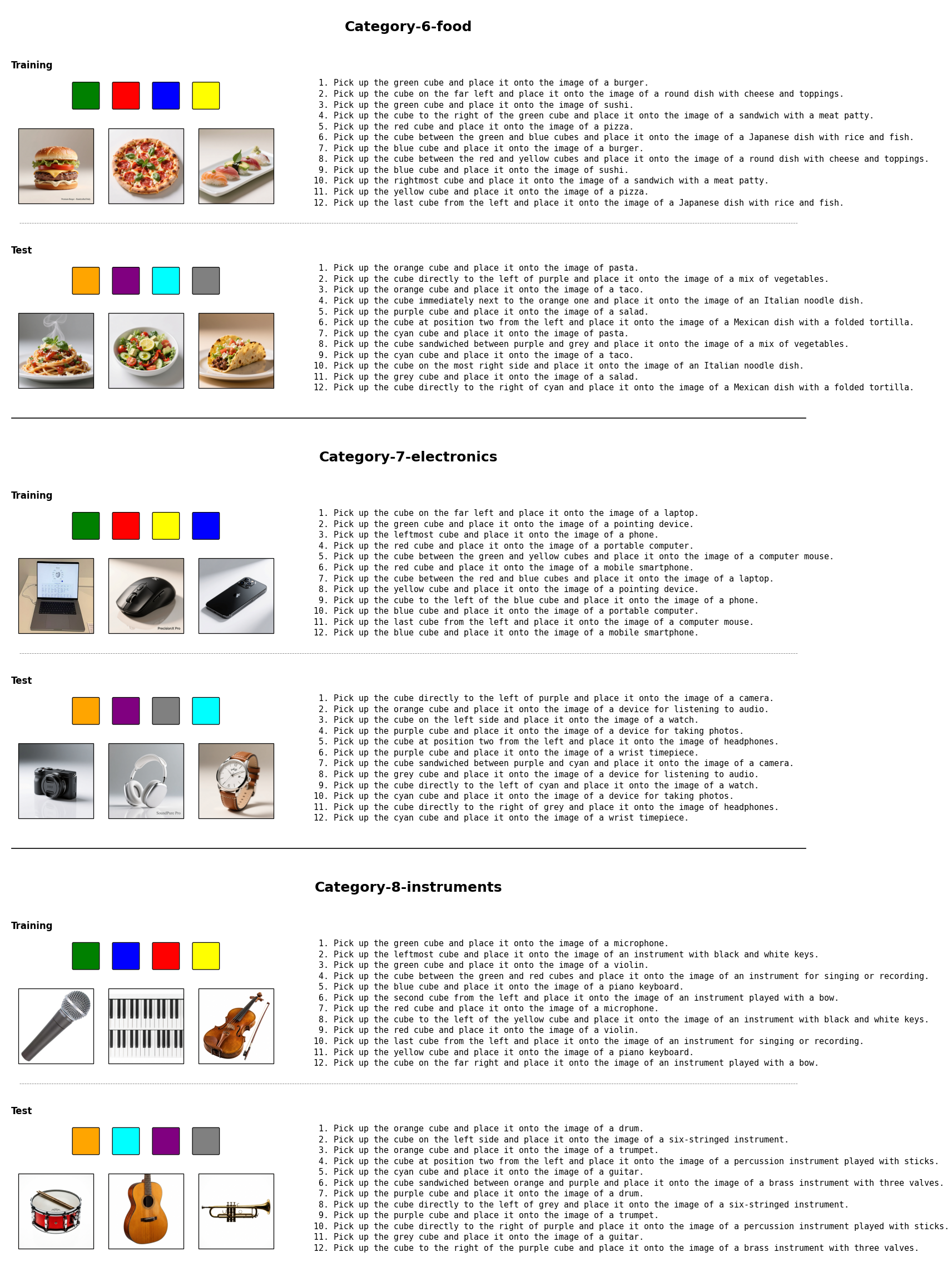}
\end{figure*}
\begin{figure*}
    \centering
    \includegraphics[width=\linewidth]{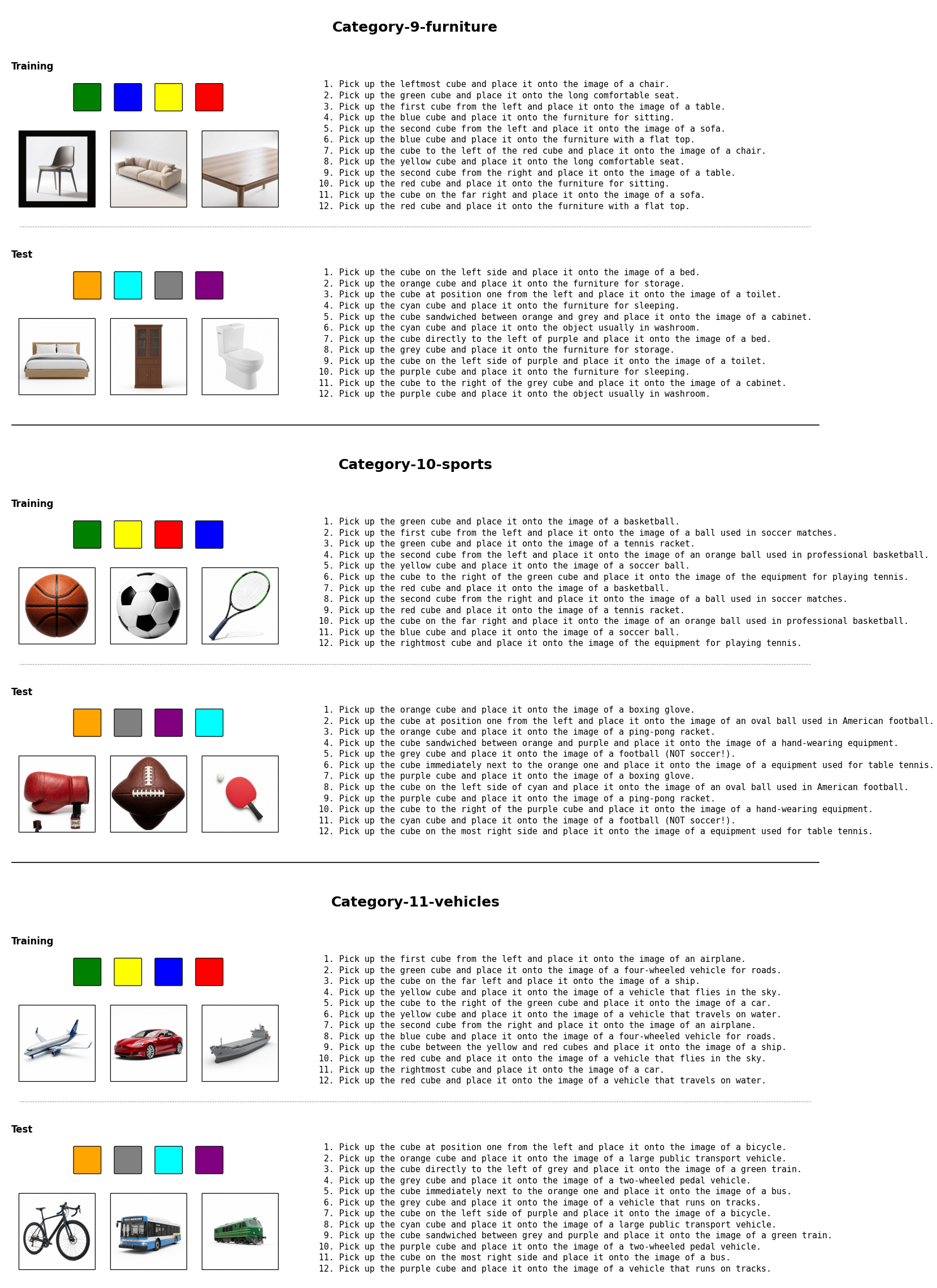}
\end{figure*}
\begin{figure*}
    \centering
    \includegraphics[width=\linewidth]{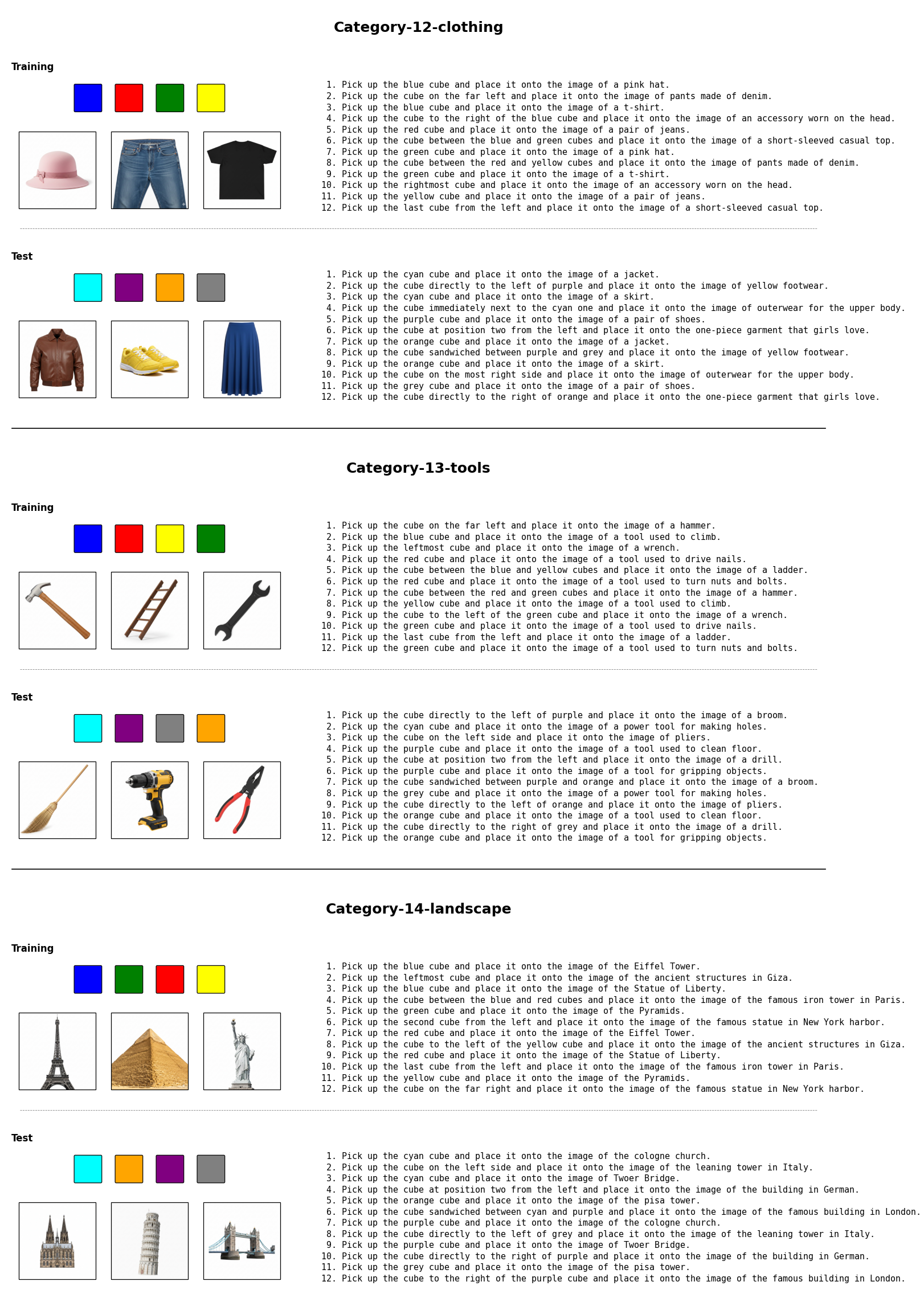}
\end{figure*}
\begin{figure*}
    \centering
    \includegraphics[width=\linewidth]{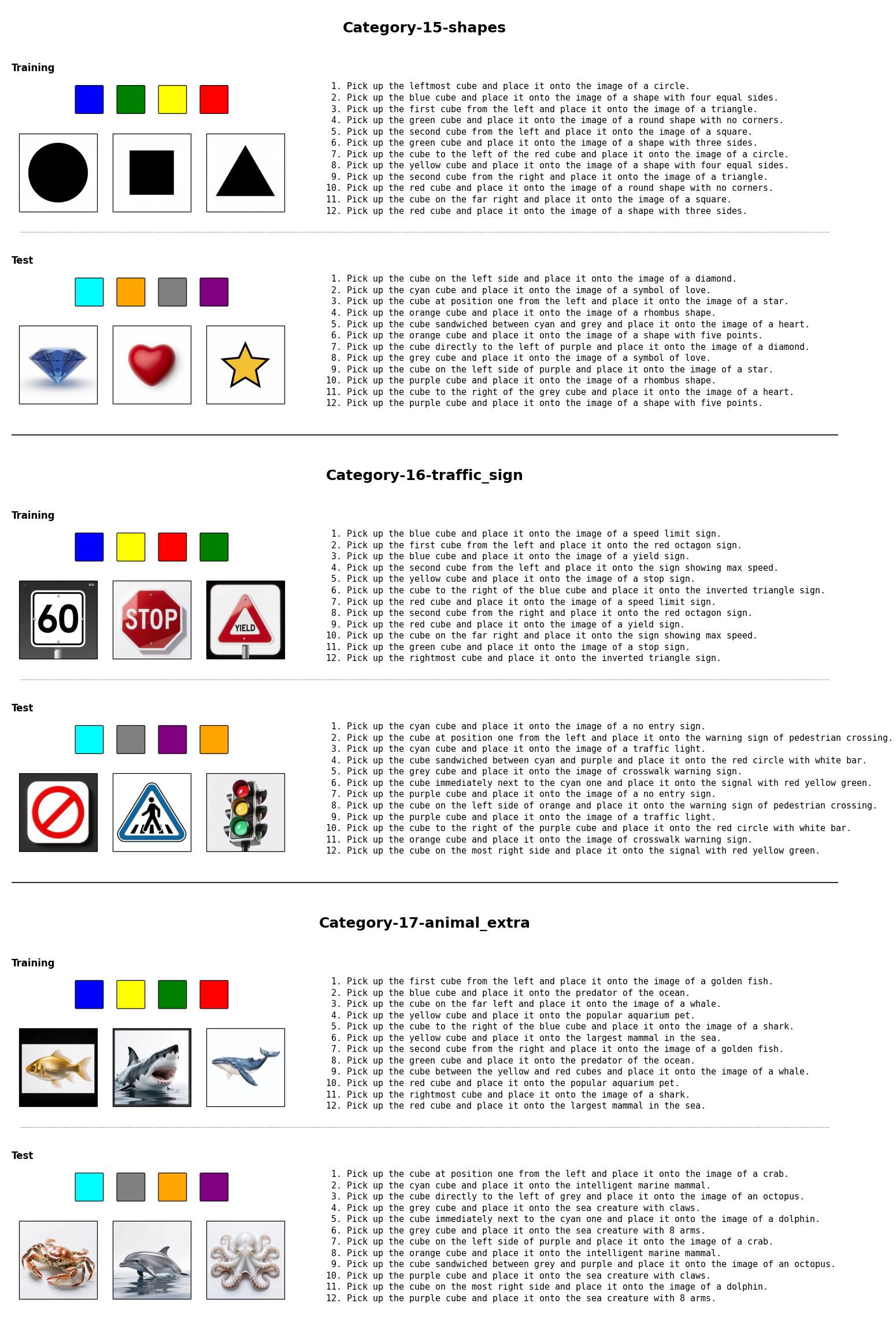}
\end{figure*}
\begin{figure*}
    \centering
    \includegraphics[width=\linewidth]{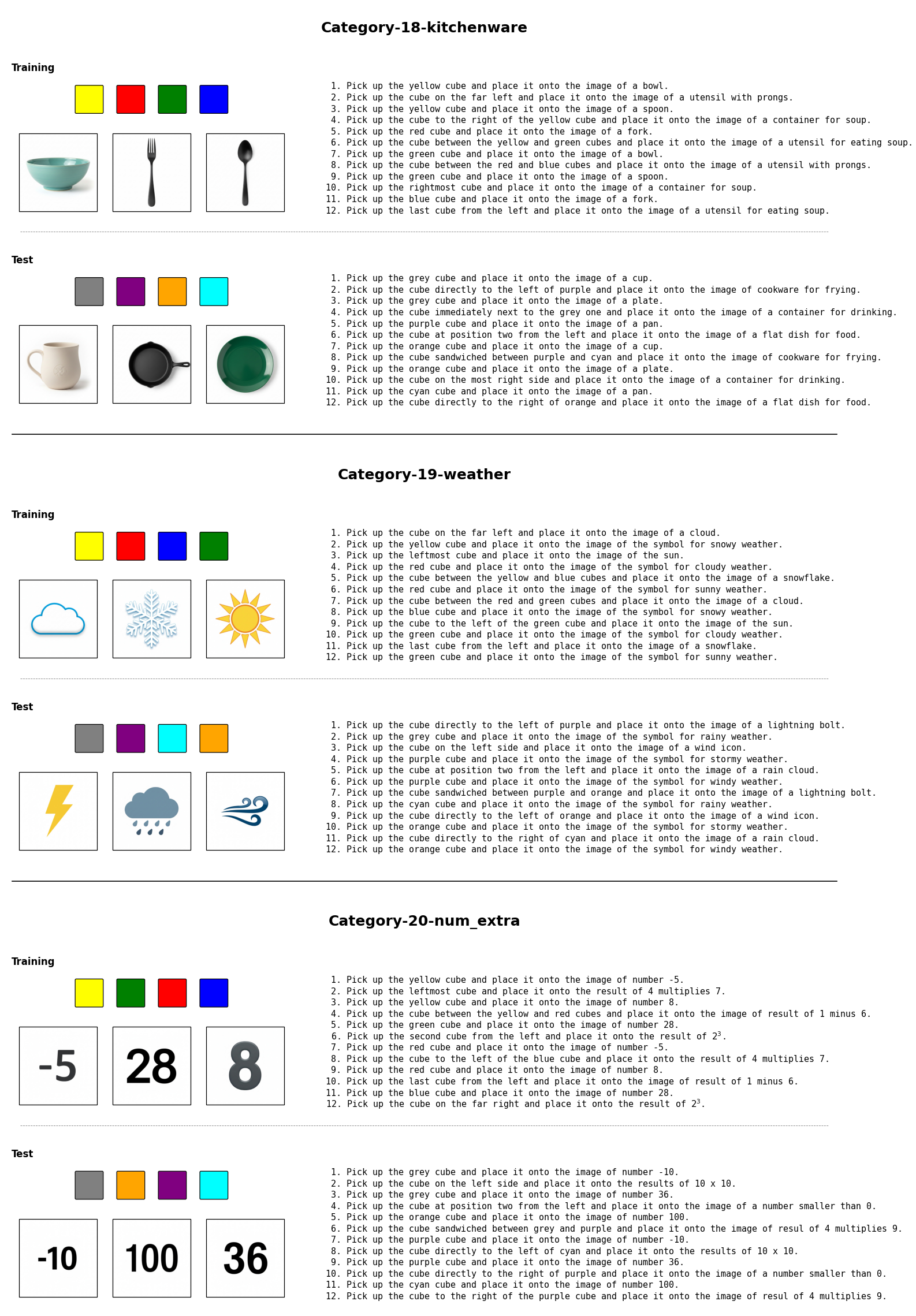}
\end{figure*}
\begin{figure*}
    \centering
    \includegraphics[width=\linewidth]{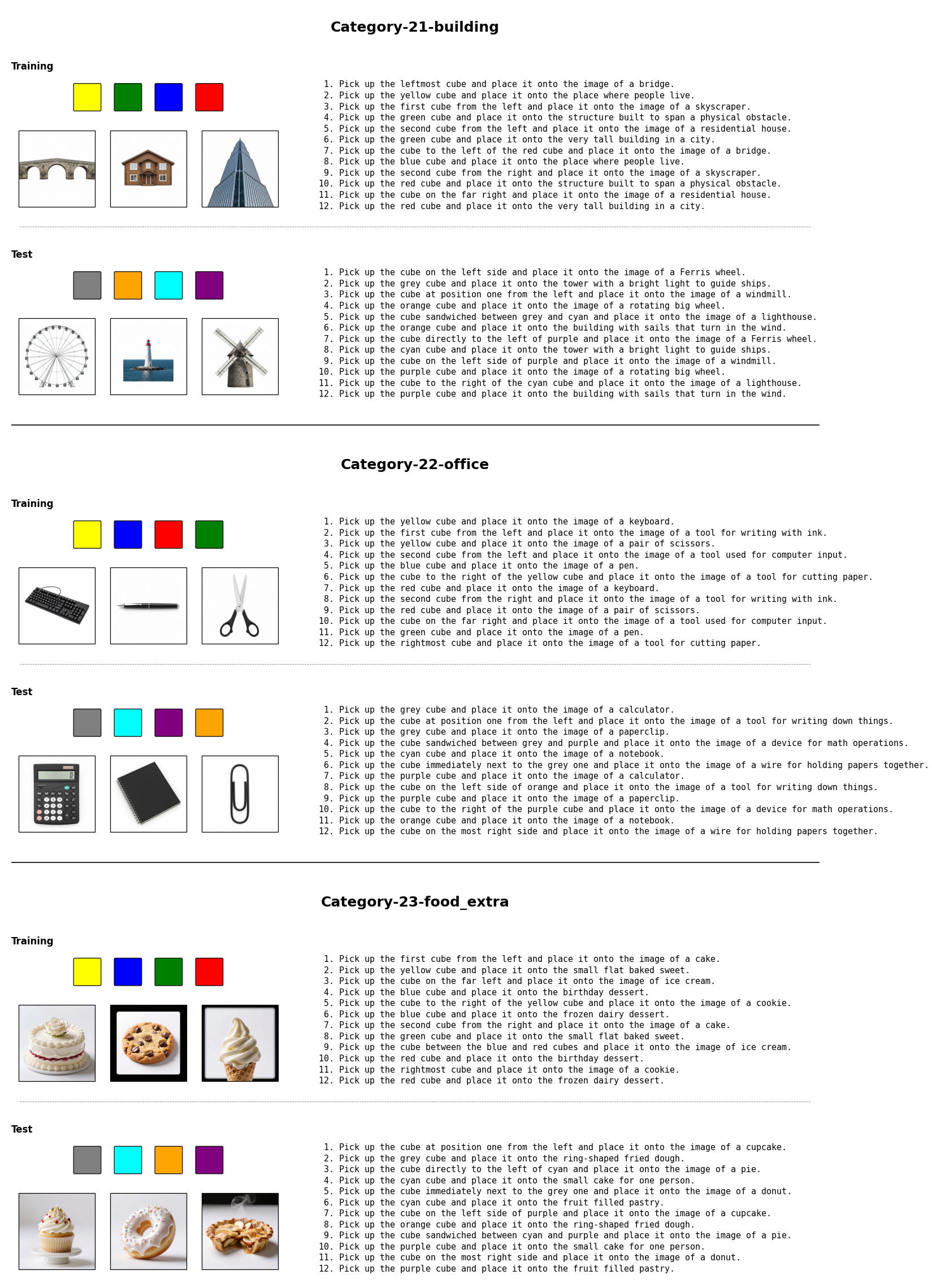}
\end{figure*}